%% file: main.tex
\documentclass[acmtog,nonacm]{acmart}
\usepackage{booktabs} 
\usepackage{amsmath}
\usepackage{enumitem}
\usepackage{graphicx}
\usepackage{subfigure}
\usepackage{multirow}
\usepackage[ruled]{algorithm2e} 

\newlength{\tempdim}

\newcommand{\img}{{\mathbf{I}}}
\newcommand{\struct}{{\mathbf{L}}}
\newcommand{\screen}{{\mathbf{S}}}

\begin{document}

\title{Screentone-Preserved Manga Retargeting}

\author{Minshan Xie}
\affiliation{%
\institution{The Chinese University of Hong Kong}
}
\email{msxie@cse.cuhk.edu.hk}
\author{Menghan Xia}
\affiliation{%
\institution{Tencent AI Lab}}
\email{menghanxyz@gmail.com}
\author{Xueting Liu}
\affiliation{%
\institution{Caritas Institute of Higher Education}}
\email{tliu@cihe.edu.hk}
\author{Tien-Tsin Wong}
\affiliation{%
\institution{The Chinese University of Hong Kong}}
\email{ttwong@cse.cuhk.edu.hk}
\renewcommand{\shortauthors}{Xie et al.}

\begin{abstract}

As a popular comic style, manga offers a unique impression by utilizing a rich set of bitonal patterns, or screentones, for illustration. However, screentones can easily be contaminated with visual-unpleasant aliasing and/or blurriness after resampling, which harms its visualization on displays of diverse resolutions.
To address this problem, we propose the first manga retargeting method that synthesizes a rescaled manga image 
while retaining the screentone in each screened region. This is a non-trivial task as accurate region-wise segmentation remains challenging. Fortunately, the rescaled manga shares the same region-wise screentone correspondences with the original manga, which enables us to simplify the screentone synthesis problem as an anchor-based proposals selection and rearrangement problem. 
Specifically, we design a novel manga sampling strategy to generate aliasing-free screentone proposals, based on hierarchical grid-based anchors that connect the correspondences between the original and the target rescaled manga. Furthermore, a Recurrent Proposal Selection Module (RPSM) is proposed to adaptively integrate these proposals for target screentone synthesis. Besides, to deal with the translation insensitivity nature of screentones, we propose a translation-invariant screentone loss to facilitate the training convergence.
Extensive qualitative and quantitative experiments are conducted to verify the effectiveness of our method, and notably compelling results are achieved compared to existing alternative techniques.
\end{abstract}



\begin{teaserfigure}
    \centering
    \subfigure[Original manga]{\includegraphics[width=.26\linewidth]{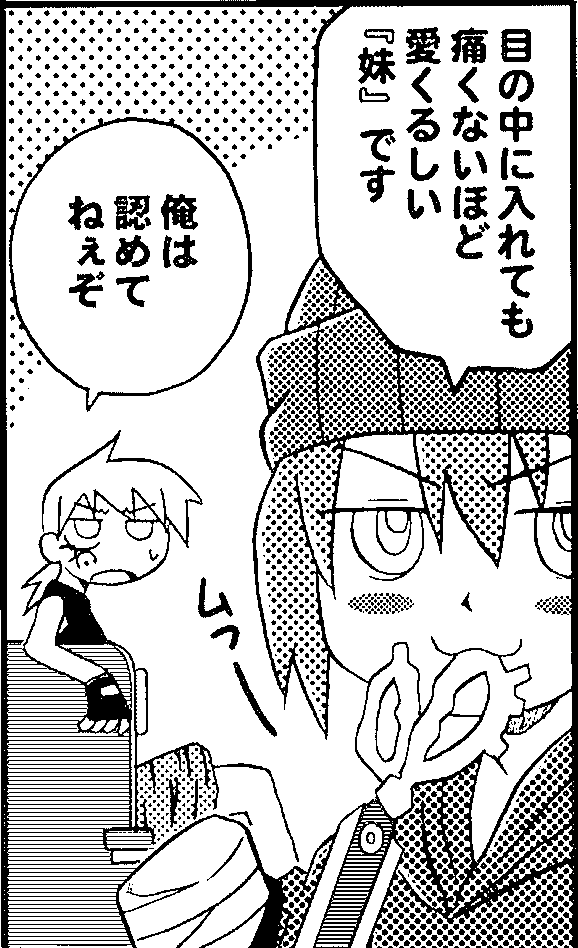}}
    \subfigure[\cite{tsubota2019synthesis}]{\includegraphics[width=.175\linewidth]{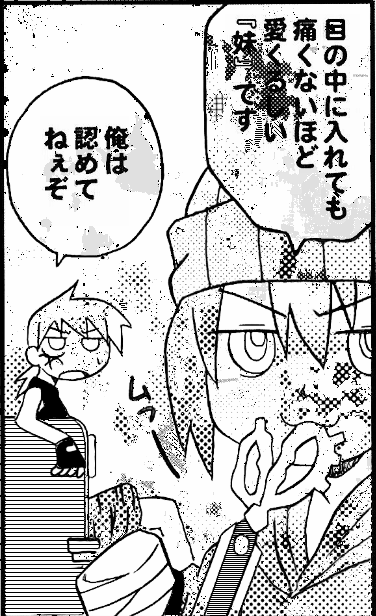}}
    \subfigure[\cite{xie2020manga}]{\includegraphics[width=.175\linewidth]{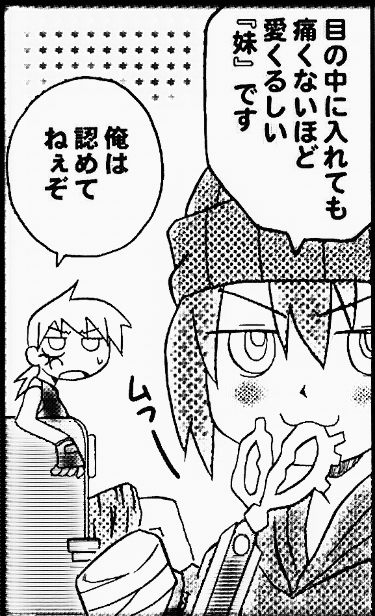}}
    \subfigure[\cite{xie2021seamless}]{\includegraphics[width=.175\linewidth]{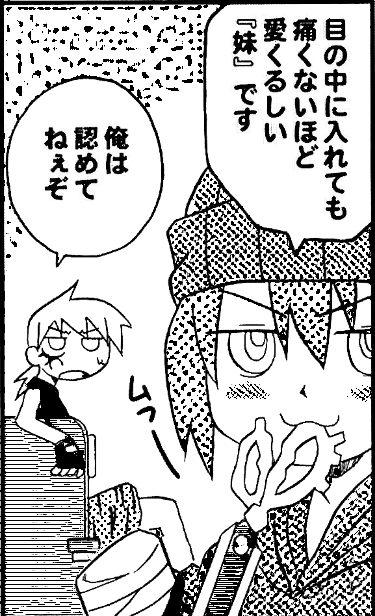}}
    \subfigure[Ours]{\includegraphics[width=.175\linewidth]{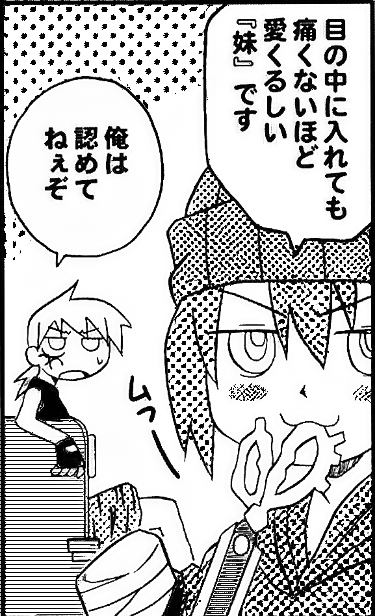}}
    \caption{The goal of manga retargeting is to retain the screentone resolution while allowing the overall manga scale to change. This avoids the  blurriness and the aliasing introduced by naive resampling. Comparing to existing applicable methods (b)-(d), our proposed retargeting model in (e) can better preserve the original screentones.}
    \label{fig:teaser}
\end{teaserfigure}

\maketitle
\input{intro}
\input{relatedwork}

\input{overview}
\input{approach}
\input{results}
\input{conclusion}

\bibliographystyle{ACM-Reference-Format}
\bibliography{main}
\end{document}

%% file: intro.tex
\section{Introduction}
\label{sec:intro}

\begin{figure}[!h]
    \centering
    \subfigure[Bitonal manga]{\includegraphics[width=.42\linewidth]{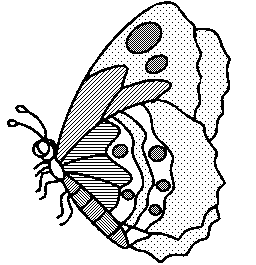}}
    \subfigure[Bilinear scaling (66\%)]{\includegraphics[width=.28\linewidth]{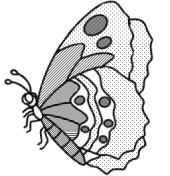}}
    \subfigure[Ideal manga]{\includegraphics[width=.28\linewidth]{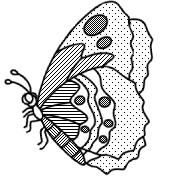}}
    \caption{Manga rescaling to fit in the reduced resolution. By bilinearly interpolation, blurry screentones and aliasing artifacts may be introduced. }
    \label{fig:senario}
\end{figure}

The popular Japanese-style comic or manga makes use of various black-and-white patterns, or {\em screentones}, to illustrate the regional tonal and texture characteristics, and better differentiate regions, as color is missing. With the popularity of mobile digital devices, such as smartphones, tablets, etc, readers are now enjoying the manga on displays of various sizes. However, to fit the display resolution, manga images are usually resampled bilinearly or bicubicly, which inevitably causes blurriness and aliasing artifacts~\cite{xie2021exploiting} on these sreentones, especially with the reduced resolution. A typical example is demonstrated in Fig.~\ref{fig:senario}.
In this scenario, a \textit{screentone-preserved manga retargeting} technique is urgently needed in order to preserve the visual fidelity of manga when presented in different resolutions.

\begin{figure}[ht]
    \centering
    \subfigure[Original manga]{\includegraphics[width=.48\linewidth]{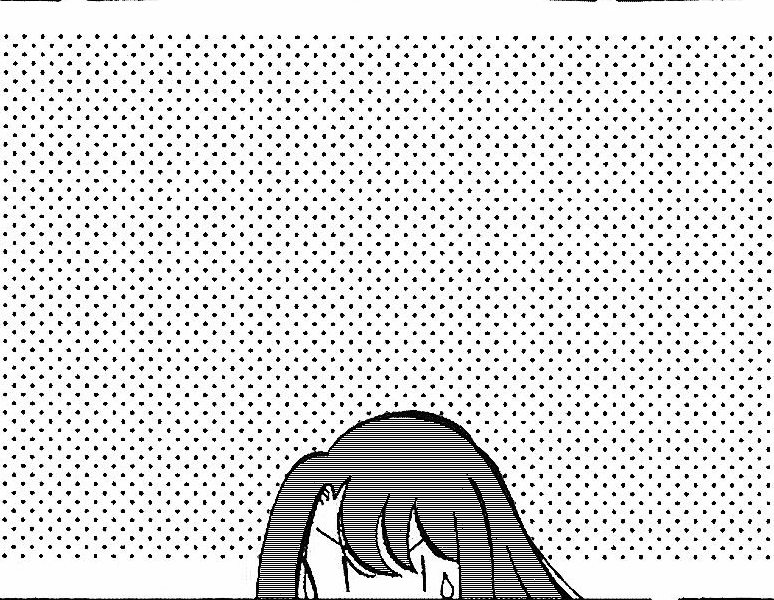}}
    \begin{minipage}[b]{0.5\linewidth}
        \subfigure[\cite{tsubota2019synthesis}]{\includegraphics[width=.48\linewidth]{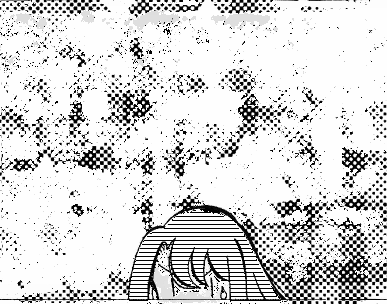}}
        \subfigure[\cite{xie2020manga}]{\includegraphics[width=.48\linewidth]{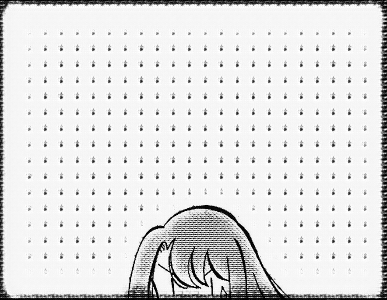}}\\
        \subfigure[\cite{xie2021seamless}]{\includegraphics[width=.48\linewidth]{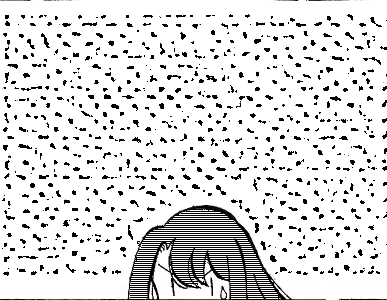}}
        \subfigure[Ours]{\includegraphics[width=.48\linewidth]{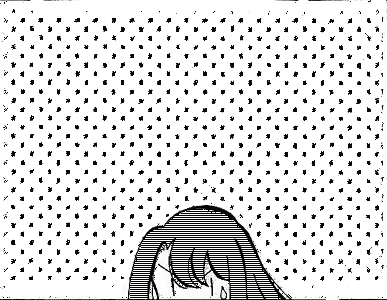}}
    \end{minipage}
    \caption{Comparison to existing methods with 50\% reduction in both width and height.}
    \label{fig:existing}
\end{figure}

\begin{figure}[ht]
    \begin{minipage}[b]{\linewidth}
        \settoheight{\tempdim}{\includegraphics[width=0.21\linewidth]{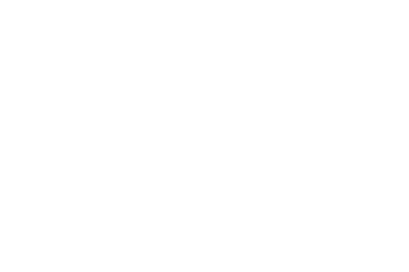}}
        \rotatebox{90}{\makebox[\tempdim]{s=80\%}}\hfil
        \includegraphics[width=0.21\linewidth]{imgs/fig1/007_orig-8.png}\hfil
        \includegraphics[width=0.21\linewidth]{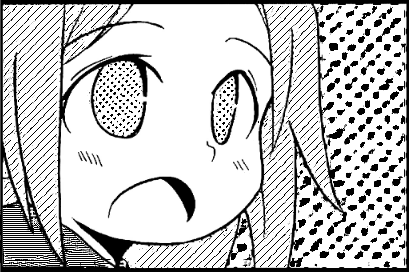}\hfil
        \includegraphics[width=0.21\linewidth]{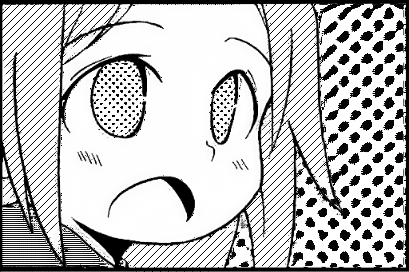}
    \end{minipage}
    \begin{minipage}[b]{\linewidth}
        \settoheight{\tempdim}{\includegraphics[width=0.265\linewidth]{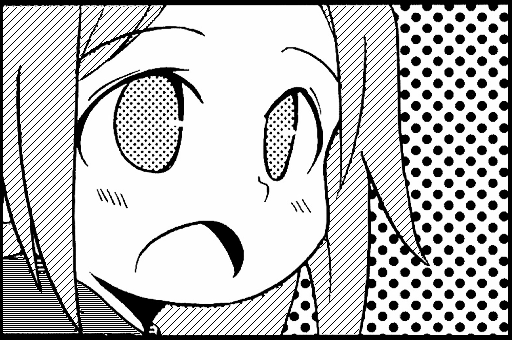}}
        \rotatebox{90}{\makebox[\tempdim]{s=100\%}}\hfil
        \includegraphics[width=0.265\linewidth]{imgs/fig1/007_orig.png}\hfil
        \includegraphics[width=0.265\linewidth]{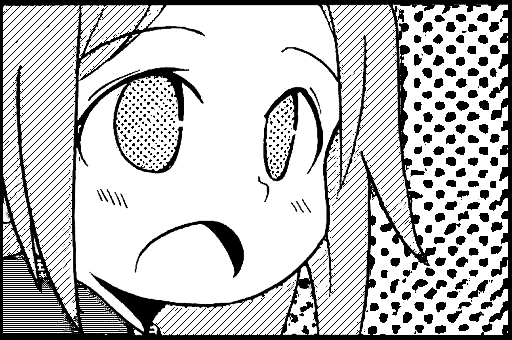}\hfil
        \includegraphics[width=0.265\linewidth]{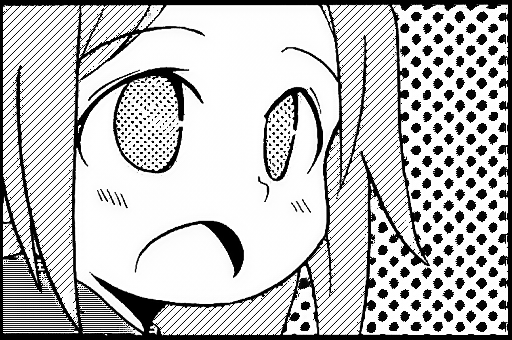}
    \end{minipage}
    \begin{minipage}[b]{\linewidth}
        \settoheight{\tempdim}{\includegraphics[width=0.32\linewidth]{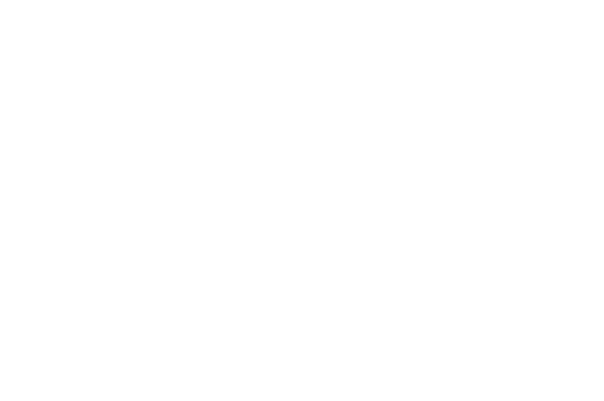}}
        \rotatebox{90}{\makebox[\tempdim]{s=120\%}}\hfil
        \includegraphics[width=0.32\linewidth]{imgs/fig1/007_orig-12.png}\hfil
        \includegraphics[width=0.32\linewidth]{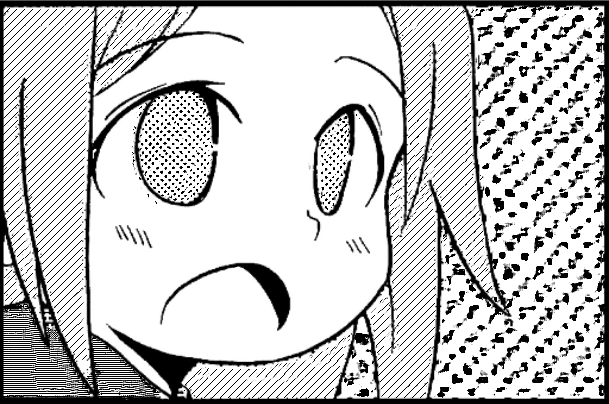}\hfil
        \includegraphics[width=0.32\linewidth]{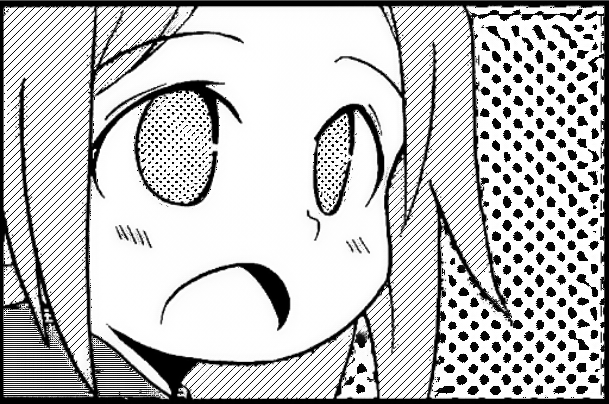}
    \end{minipage}
    \medskip
    \hspace{0.25\baselineskip}\hfil
    \makebox[0.3\linewidth]{(a) Original manga}\hfil
    \makebox[0.3\linewidth]{(b) \cite{xie2021seamless}}\hfil
    \makebox[0.3\linewidth]{(c) Ours}
    \caption{Manga inpainting versus our method in various scales.  }
    \label{fig:mangainpaint}
\end{figure}

Unlike the existing natural image retargeting techniques~\cite{lei2017depth,dong2015image}  aiming at preserving the prominent content when the image aspect ratio changes, our manga retargeting focuses on preserving the screentones when the overall image resolution changes.
In other words, we want to retain the screentone resolution within the screened region while the overall structure is allowed to rescale (compare Fig.~\ref{fig:senario}(a) to (c)).
Apparently, applying existing natural image retargeting methods to manga will damage the screentone patterns. Matsui et al.~\shortcite{matsui2011interactive} made attempts to generate manga for other aspect ratios, but they only consider line drawing while the more challenging screentone is ignored.
An intuitive solution to generate retargeted manga is to first segment the screentones regions and then applied screentone patches to the corresponding regions~\cite{yao2016manga,tsubota2019synthesis}. However, these methods can only handle screentones in a predefined set and may not generate consistent screentone within the same region, as demonstrated in Fig.~\ref{fig:existing}(b). 
Another possible solution is to adopt an interpolative representation, such as the ScreenVAE map generated by Screentone Variational AutoEncoder~\cite{xie2020manga}. This representation can be resampled and used to reconstruct screentones that resemble the original ones. However,  the reconstructed screentones (Fig.~\ref{fig:existing}(c)) may not match well with the originals. Manga inpainting technique~\cite{xie2021seamless} may also be employed for retargeting purpose, but their model is tailored for inpainting manga (both screentone and outline) at a single scale. Retargeting at arbitrary scales may fail their model, as demonstrated by their inconsistent screentones in changed scales Fig.~\ref{fig:mangainpaint}. 

\begin{figure}[!t]
    \centering
    \includegraphics[width=.95\linewidth]{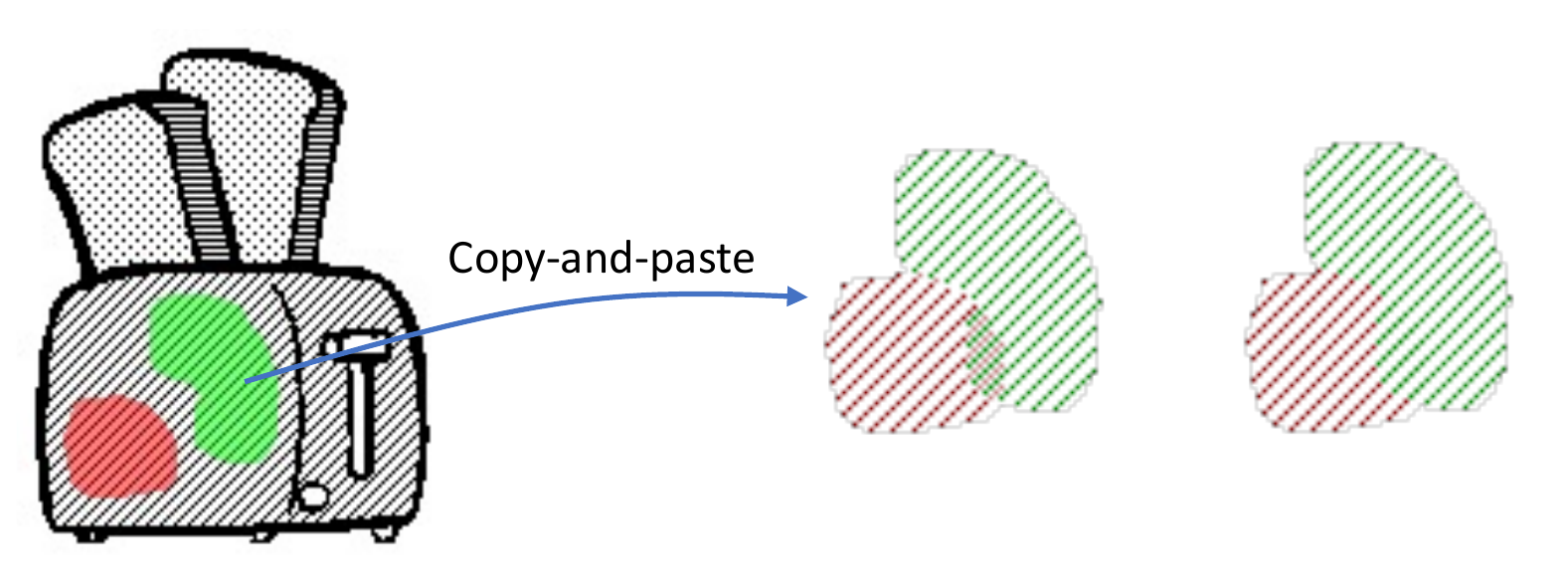}
    \medskip
    \makebox[0.3\linewidth]{(a) Input}\hfil
    \hspace{3\baselineskip}\hfil
    \makebox[0.25\linewidth]{(b) Poor-aligned}\hfil
    \makebox[0.25\linewidth]{(c) Well-aligned}
    \caption{We patch-wisely copy-and-paste the screentone from the input image to the retargeted image with well-alignment. }
    \label{fig:intuition}
\end{figure}

In this paper, we propose a learning-based model for screentone-preserved manga retargeting. As screentones tolerate no interpolation, we aim at retaining the screentone in each shrinked (enlarged) region in a shrinked (enlarged) manga. 
Fig.~\ref{fig:intuition} illustrates the intuition of our model. Intuitively speaking, we patch-wisely copy-and-paste the screentone from the input image to form a shrinked (enlarged) screened region in the shrinked (enlarged) manga. The core challenge is to ensure the screentone patches are visually well-aligned (Fig.~\ref{fig:intuition}(c)). Previous learning-based method~\cite{xie2021seamless} suffers from the poor alignment problem (Fig.~\ref{fig:intuition}(b)) and cannot generate satisfactory results, as demonstrated by the misalignment seams within the generated screentone especially the shrinked and enlarged mangas in Fig.~\ref{fig:mangainpaint}(b).

To achieve the goal, we adopt the semantic representation, consisting of structural line and ScreenVAE map, to represent the retargeted manga image, similar to~\cite{xie2021seamless}. 
However, when the scale of the manga is changed, such semantic representation is not sufficient to accurately reconstruct the original screentones. The ScreenVAE map is not helpful to align screentone during a patch-wise copy-and-paste process, as all ScreenVAE values in the same screened region are equal. This means we need to utilize the input screentone in spatial domain to align the patches during our screentone synthesis.
To determine patches and align them, 
we propose a hierarchical anchor-based proposal sampling scheme to build the correspondences between the original and the retargeted screentone layouts. These proposals are intuitively our ``patches for copy-and-paste'' and they are aligned in this module. Next, we design a Recurrent Proposal Selection Module (RPSM) to fuse (``paste'') these proposals one by one 
and, simultaneously, hold an accumulated confidence map to avoid the regions already fused with high-confidence proposal features to be fused again.
To make it works, we also propose a translation-invariant screentone loss to allow the generated screentones to match one feasible solution, which is practically useful to guarantee sharp screentone patterns.

To evaluate our method, extensive experiments are conducted, including qualitative and quantitative results of synthetic data. Convincing results are obtained in terms of visual quality and quantitative evaluation. Also, we apply our method on various real-world manga images to generate manga with diverse resolutions, and visually plausible results are obtained. Our contributions can be summarized as follows:
\begin{itemize}
    \item We propose the first manga retargeting method, which generates manga with structures resampled while screentones maintaining at the original scale. 
    \item We propose a Manga Retargeting Network with sampled proposals to select and rearrange appropriate ones for screentone synthesis. 
    \item We propose a translation-invariant screentone loss to tolerate the screentone misalignment and enable the learning from multiple solutions. 
\end{itemize}

%% file: relatedwork.tex
\section{Related work}

\subsection{Image Retargeting}
Retargeting is the process of adapting an image from one screen resolution to another, often with the change of the aspect ratio as well \cite{shamir2009visual}. 
In real-time applications, global visual effects of the image can be preserved using scaling operators when interpolation scaling methods are employed. But, these methods may produce distortion of important contents. 
Plenty of content-aware retargeting approaches have been proposed to improve the visual quality of the resized images \cite{lei2017depth,dong2015image}. The existing content-aware image retargeting algorithms can be mainly categorized into two types: discrete approaches and continuous approaches. 

In the discrete approaches, an input image is resized by cropping or seam carving. Cropping-based methods \cite{liu2006video,suh2003automatic} select an optimal rectangle from the image and remove the content outside the rectangle. Still, they may lead to noticeable information loss and not perform well with images containing multiple salient objects. Seam carving \cite{avidan2007seam,shamir2009visual} is another typical discrete approach, which achieves image resizing by removing or inserting one seam with the least energy each time until reaching the target size. However, it will cause obvious distortion if too many seams passing through important objects are involved. 
Rubinstein et al.\shortcite{rubinstein2009multi} present a multi-operator algorithm and get more pleasing results than other single operators by combining cropping, linear scaling, and seam carving. 
Pritch et al.\shortcite{pritch2009shift} regards image retargeting as geometric rearrangement of images and discretely removes repeated patterns in homogenous regions instead of scaling and stretching images. 
Although these approaches can generate pleasing results for many cases, it should be noted that all the above discrete methods may lead to noticeably jagged edges and discontinuous artifacts in image objects. 

Instead of eliminating redundant contents directly, continuous methods optimize a mapping or warping using several deformation and smoothness constraints. 
Liu et al.\shortcite{liu2005automatic} propose a non-linear warping scheme to preserve important content. However, contents outside the region of interest may suffer from significant distortions. 
Wolf et al.\shortcite{wolf2007non} propose to reduce distortion by merging less important pixels. However, the distortion still propagates along the resizing direction. 
To refine the distortion propagation, Wang et al. \shortcite{wang2008optimized} propose an optimized scale-and-stretch approach which divides an image into uniform mesh grand and resizes the image by iteratively computing optimal local scaling factors of local regions. But this method may lead to inconsistent deformation for objects occupying several mesh. 
To ease this problem, some approaches \cite{zhang2009shape,guo2009image,jin2010nonhomogeneous} propose to combine saliency information to protect salient objects. These methods perform well on preserving the shape of local objects while inconsistent distortions may still occur on structure lines. 

Deep learning techniques have achieved astonishing advances in recent years and also promoted the development of image retargeting. Liu et al.\shortcite{liu2018composing} use deep neural networks to extract semantic components, which are then fused with the original image to generate the target image. Cho et al.\shortcite{cho2017weakly} first applied deep learning in image retargeting in an end-to-end manner. A weakly- and self-supervised deep convolutional neural network (WSSDCNN) is proposed to predict attentive shift maps which are used to warp the input image into an image of target size. However, these methods require image-level annotations when training. Tan et al.\shortcite{tan2019cycle} proposed a deep cyclic image retargeting approach, called Cycle-IR, without relying on any explicit user annotations, but it cannot perform well when important areas are too large or scattered or with low contrast. DeepIR\cite{lin2019deepir} first constructs the semantic structures with a deep neural network and then applies uniform resampling in feature level to preserve these structures in the retargeted image, while it may over squeeze the important regions or over maintain the less important ones. 
All above image retargeting methods cannot handle manga as they all involve scaling operations on image contents which will destroy the appearance of screentones.


\begin{figure*}[!t]
    \centering
    \includegraphics[width=\linewidth]{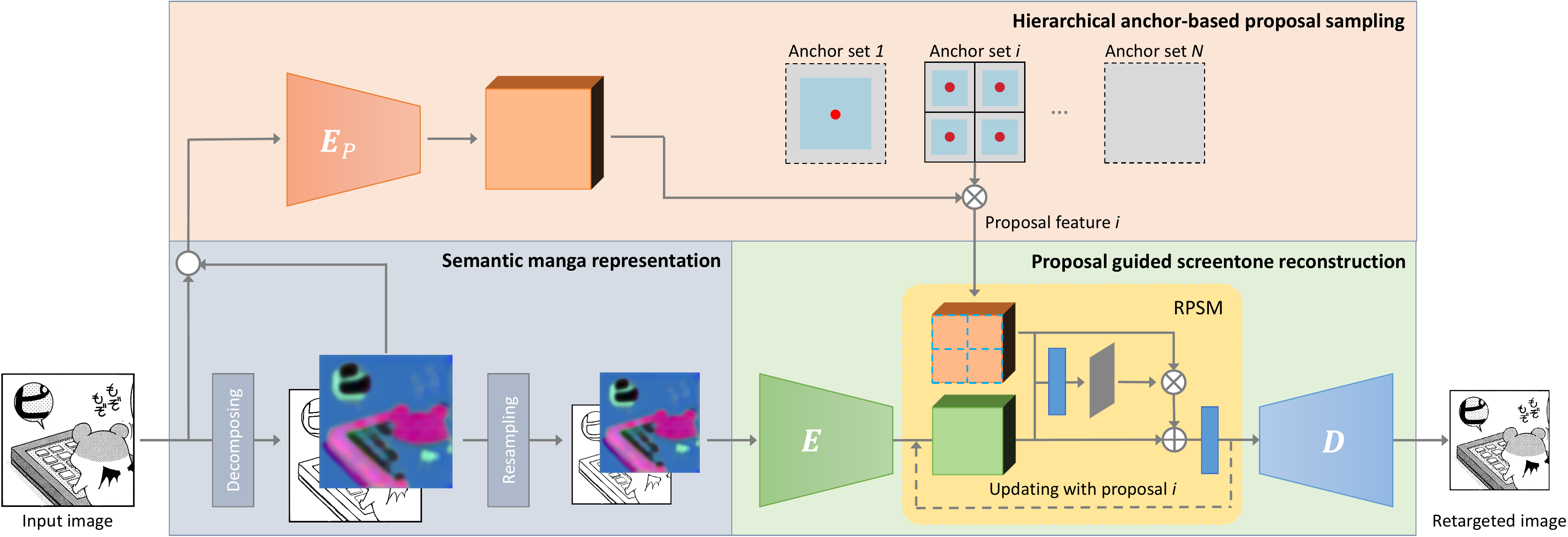}
    \caption{Overview diagram of our manga retargeting system. Given a high-quality manga image $\img$, we first decompose it into structural line $\mathbf{L}$ and ScreenVAE map $\screen$, which are resampled to the target size via bilinear interpolation, and then fed to the screentone synthesis network. Meanwhile, the input manga $\img$ and its ScreenVAE map are fed to another feature extractor $\mathbf{E}_p$ that offers screentone features for hierarchical anchor-based proposal sampling. Then, the proposal screentone features are integrated by the Recurrent Proposal Selection Module (RPSM) in a progressive manner, and the fused features are finally used to reconstruct the retargeted manga image $\tilde{\img}$.}
    \label{fig:overview}
\end{figure*}

\subsection{Screentone Extraction and Manga Screening}
Some attempts have been made to extract screentones from manga and generate new screentone patches.
Yao et al.\shortcite{yao2016manga} considered how to extract screentones from manga by modeling three specific texture primitives, dots, stripes, and grids. These primitives can be used to generate larger patches after completion. 
Tsubota et al.\shortcite{tsubota2019synthesis} synthesize manga images by generating pixel-wise screentone class labels and further laying the corresponding screentones from the database. However, these methods are highly dependent on the predefined screentone set and may not generate the original screentones, even with tone-varying patterns. 
Xie et al.\shortcite{xie2020manga} proposed a Screentone Variational Autoencoder (ScreenVAE) model to build a bidirectional mapping of screened manga  to an intermediate latent space. The intermediate representation is interpolative and can reconstruct the original screentone. But, the reconstructed screentones are somewhat different from the original version due to the dataset bias \cite{xie2021seamless}, especially with large-scale patterns. 
Xie et al.\shortcite{xie2021seamless} use the previous representation as a hint and further generate consistent screentones by borrowing screentones from the original manga image. However, this method is tailored for inpainting manga at a single scale, and they may fail when retargeting manga at arbitrary scales. 
In contrast, we use multiple proposals to ease the generation of screentones and generate retargeted manga by selecting and aligning appropriate proposals.


%% file: overview.tex
\section{Overview}
\label{sec:overview}


Given a high-quality manga image $\img$ of size $H\times W$, we aim at synthesizing a retargeted version $\hat{\img}$ of size $kH \times kW$ (where $k \in \mathbb{R}_+$), which has the structure lines resampled accordingly while preserves the screentones at the original scale. As screentone is vulnerable to interpolation, we follow the practice of~\cite{xie2021seamless} and adopt semantic representation of manga images, including structure line and ScreentoneVAE map, so that we can approximately represent the desired manga by interpolating in this domain. Anyhow, such representation is not informative enough to reconstruct the original screentones, possibly because of the dataset bias learned by the ScreentoneVAE map generator~\cite{xie2020manga}. Regarding this, we additionally exploit the original screentone to extract proposals based on a hierarchical anchor sampling scheme, which provides effective guidance to promote the screentone reconstruction fidelity.

The overview diagram of our system is illustrated in Figure~\ref{fig:overview}, including three functional blocks.
Firstly, the input manga image $\img$ is decomposed into a structural line map $\mathbf{L}$ and a screentone image $\img^s$ through~\cite{li2017deep}. The screentone image is further encoded as the ScreenVAE map $\screen$ with smooth values within each screentone region ~\cite{xie2020manga}. $\mathbf{L}$ and $\screen$ together forms the semantic representation of $\img$.
Then, we feed the resampled semantic representation $\tilde{\mathbf{L}}$ and $\tilde{\screen}$ to the screentone reconstruction network $G$. To supplement screentone details, a special feature extractor $E_p$ is employed to encode region-wise screentone features $\mathbf{F}_s$ from the input manga image $\img$ and $\screen$. $\mathbf{F}_s$ will be integrated with the bottleneck features $\mathbf{F}_b$ of $G$ through the Recurrent Proposal Selection Module (RPSM).
Note that, $\mathbf{F}_s$ and $\mathbf{F}_b$ are under different resolutions and directly resampling $\mathbf{F}_s$ may ruin the embedded high-frequency screentone details. So, to construct proposals of the same resolution, we propose to regularly scatter a grid of anchors $\mathcal{A}=\{\mathbf{a}_{i,j}\}^{I,J}$ and crop the local patches of $\mathbf{F}_s$ that are centralized on these anchors. Particularly, we construct multiple groups of such proposals by changing the anchor-grid interval (i.e. the anchor numbers), which offer more rich information for screentone reconstruction.
Considering these proposal patches may not fully match the target regions, we propose Recurrent Proposal Selection Module (RPSM) to progressively select appropriate proposals, and the fused features are exploited to generate the resultant manga image.

\begin{figure}[!h]
    \centering
    \subfigure[Screentone 1]
    {\includegraphics[width=.235\linewidth]{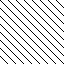}}
    \subfigure[Screentone 2]
    {\includegraphics[width=.235\linewidth]{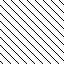}}
    \subfigure[Screentone 3]
    {\includegraphics[width=.235\linewidth]{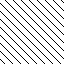}}
    \subfigure[Overlapped screentone]
    {\includegraphics[width=.235\linewidth]{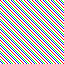}}
    \caption{The three patterns are the same screentone with translation.}
    \label{fig:misalign}
\end{figure}

As consisting of periodic or aperiodic discrete patterns, screentones are perceptually insensitive to pattern translation, as demonstrated in Fig.~\ref{fig:misalign}. These characteristic poses challenge to the training of screentone synthesis because one given ground-truth actually corresponds to lots of other variants of the same essence.
Both pixel-wise measurement or using window averaged features will cause blurry results as a kind of average effect.
To tackle this issue, we proposed a Translation-Invariant Screentone Loss that allows the generated screentones to match one possible solution within a tolerance window. It comes out to be essential for generating discrete high-frequency screentones.
The detailed model architectures and loss functions are described in Section \ref{sec:approach}.

%% file: approach.tex
\section{Our Approach}
\label{sec:approach}

In this section, we first describe the network architecture logically, and then introduce two key technical designs in detail. After that, the loss function will be presented.

\subsection{Network Architecture}
\label{subsec:network_architecture}

Our manga retargeting network contains two branches, namely the screentone reconstruction branch $G$ and the screentone encoder branch $\mathbf{E}_p$.
The screentone reconstruction network $G$ adopts an encoder-decoder structure, implemented with a series of convolution layers and dilated residual blocks\cite{yu2017dilated}. It takes the resampled semantic components (structural line map $\tilde{\mathbf{L}}$ and ScreenVAE map $\tilde{\screen}$) as input and generate the screentone-preserved manga image $\tilde{\img}$ under the guidance of the original screentone proposals. Particularly, the screentone proposals are fused in the bottleneck of $G$ and through the Recurrent Proposal Selection Module (RPSM) that will be expatiated in section ~\ref{subsec:proposal_sampling}.

Since the semantic representation is not informative enough to reconstruct the original screentone precisely, we employ a special screentone encoder $E_p$ to extract screentone appearance features from the original manga image $\img$ and its ScreenVAE map $\screen$, which will be sampled under a hierarchical anchor-based proposal sampling scheme (detailed in Section~\ref{subsec:proposal_sampling}) to provide guidance for $G$. The screentone encoder $E_p$ shares the same architecture with the encoder of $G$, and the decoder $D$ is with symmetric architecture with the encoder $E$. Skip connections are adopted between $E$ and $D$ to preserve the spatial information of regions.
The detailed network architectures are provided in the supplementary material.

\subsection{Hierarchical Proposal based Recurrent Fusion}
\label{subsec:proposal_sampling}

\begin{figure}[!t]
    \centering
    \subfigure[Reduce resolution]{\includegraphics[width=.49\linewidth]{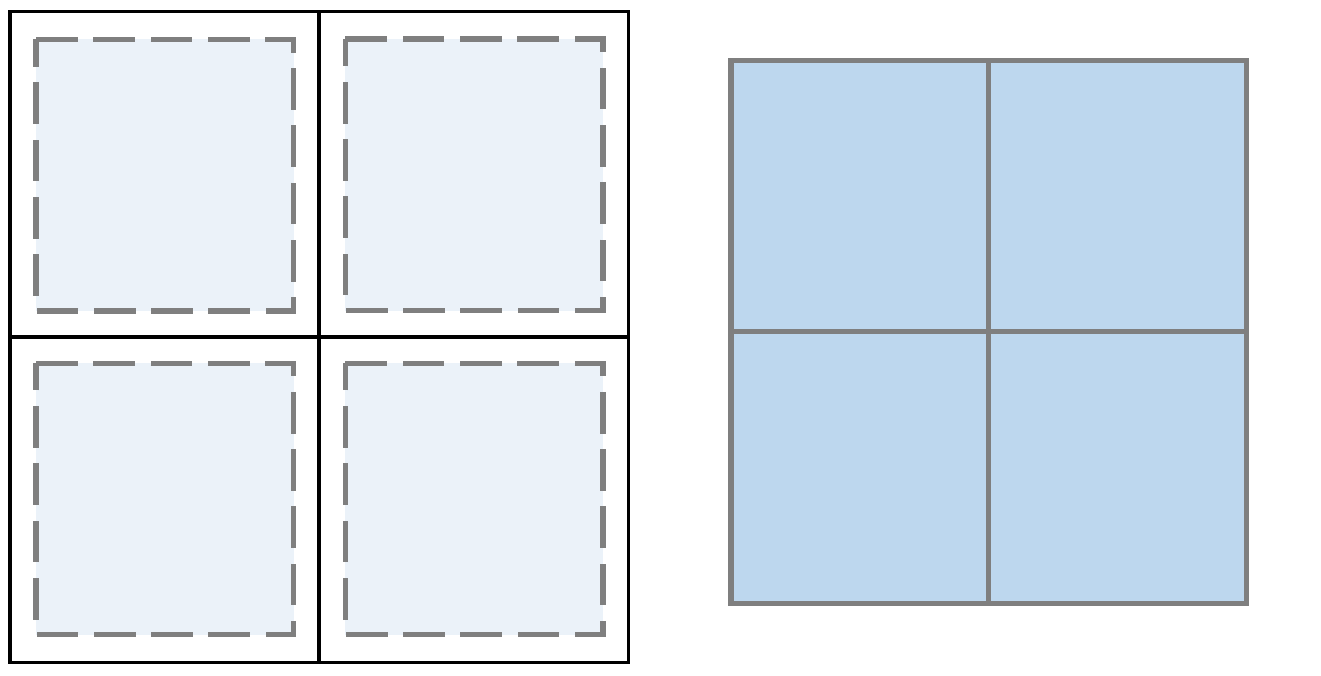}}
    \subfigure[Enlarge resolution]{\includegraphics[width=.49\linewidth]{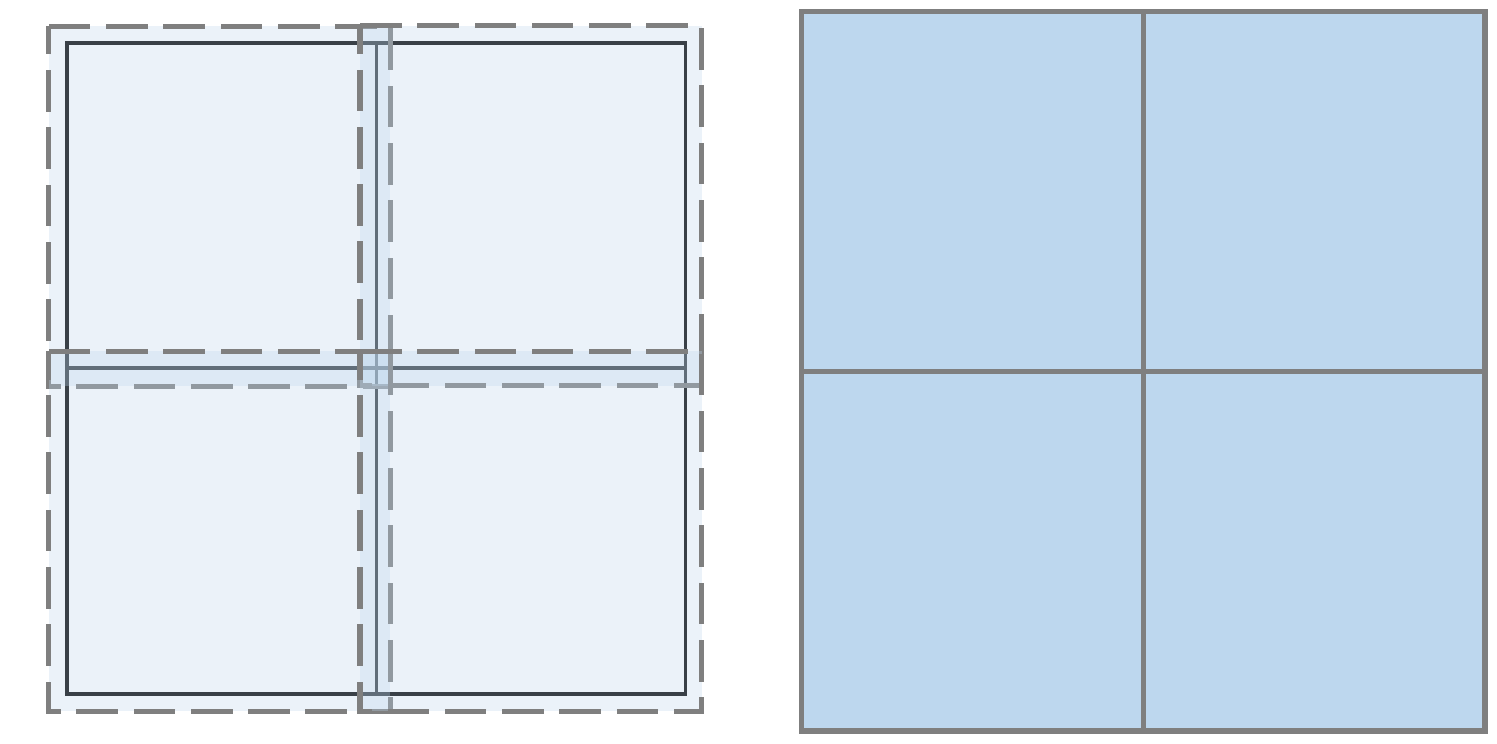}}
    \caption{Illustration of the grid-anchor-based proposal sampling. Local feature patch around each anchor is cropped, and all the patches are combined to form a feature proposal. }
    \label{fig:patches}
\end{figure}

\paragraph{Hierarchical anchor-based proposal sampling}
Given the screentone feature $\mathbf{F}_s=E_p(\img,\screen)$ of size $H_F \times W_F$, we can not fuse it with the backbone features $\mathbf{F}_b=E(\tilde{\mathbf{L}},\tilde{\screen})$ directly, because of the misalignment caused by inconsistent resolutions. Furthermore, resampling $\mathbf{F}_s$ to the same resolution of $\mathbf{F}_b$ (i.e. $kH_F \times kW_F$) may destroy the high-frequency pattern information embedded in. Instead, to ensure the intactness of local patches, we regularly scatter a $I \times J$ grid of anchors $\mathcal{A}=\{\mathbf{a}_{i,j}\}^{I,J}$ on $\mathbf{F}_s$ and sample the local patches around them. Specifically, for an arbitrary anchor $\mathbf{a}_{i,j}$, the sampled feature patch can be denoted as $\mathbf{F}_s^{i,j}=P(\mathbf{F}_s,\mathbf{a}_{i,j})$, where $P(\cdot)$ denotes a cropping function that takes the patch centralized on $\mathbf{a}_{i,j}$ and sized $\frac{kH_F}{I} \times \frac{kW_F}{J}$. For the case of $k>1$, where some cropping patches may cover regions beyond $\mathbf{F}_s$, we simply extend $\mathbf{F}_s$ by padding empty value. Then, we can combine these grid-anchor based local patches $\{\mathbf{F}_s^{i,j}\}^{I,J}$ to get a screentone feature proposal $\hat{\mathbf{F}}_s$, apparently with the same size as $\mathbf{F}_b$. Fig.~\ref{fig:patches} visualizes the sampling process, where an example with $2 \times 2$ grid of anchors is used.
Obviously, the proposal features may still have some spatial misalignment with respect to the desired screentone layout, and the regions closer to anchors shall have a better match with the proposal features. It seemingly indicates that the proposal based on denser anchors is more accurate, but it is not always the case because denser anchors mean smaller local patch size, which inevitably destroys the feature intactness for those large regions. Regarding this, we construct multiple groups of such proposals by changing the anchor-grid interval (i.e., the anchor numbers $I \times J$), which together provide more informative guidance for screentone reconstruction.

\begin{figure}[h]
    \centering
    \subfigure[Original image]{\includegraphics[width=.32\linewidth]{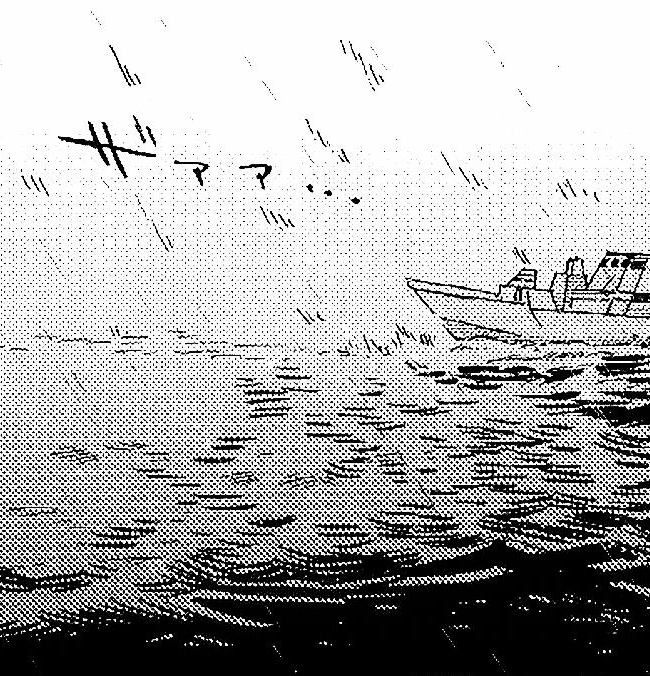}}
    \subfigure[w/o recurrent scheme]{\includegraphics[width=.32\linewidth]{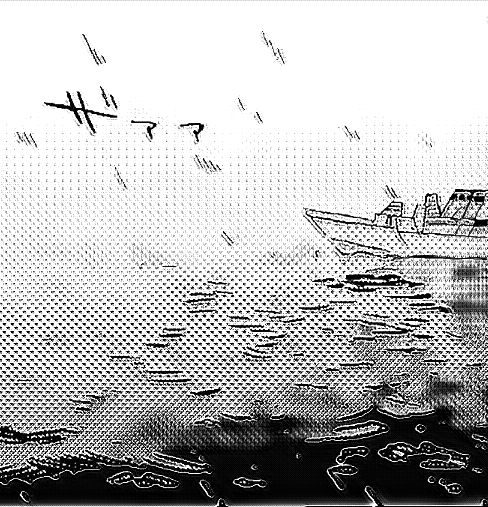}}
    \subfigure[Ours]{\includegraphics[width=.32\linewidth]{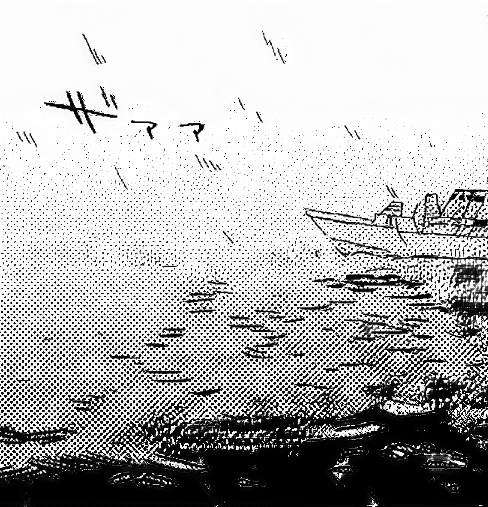}}
    \caption{Results without a recurrent scheme will be confused by multiple proposals and cannot preserve the original appearance. }
    \label{fig:progressive}
\end{figure}

\paragraph{Recurrent selection of hierarchical proposals}
Once the multiple proposals of screentone features are constructed, the screentone reconstruction network $G$ takes them as guidance to reconstruct the region-wise screentones according to the semantic information represented by $\mathbf{L}$ and $\screen$. 
A naive solution is to simply concatenate all these proposal feature sets $\{\hat{\mathbf{F}}_s^l\}_{l=0}^L$ and the backbone feature $\mathbf{F}_b$ which are then fed to decoder $D$. However, our experiment shows that such feature concatenation causes blurry reconstruction or even artifacts because the network might be confused by the case that multiple proposals provide equally-qualified solutions for certain regions. A visual example is compared in Fig.~\ref{fig:progressive}.
To address this issue, we propose an explicit proposal feature selection module, namely Recurrent Proposal Selection Module (RPSM), whose working diagram is illustrated in Fig.~\ref{fig:overview}. The primary motivation is to force the block only to choose one feasible solution from the provided proposals. We select proposal features by computing spatial attention maps in a recurrent manner. Particularly, we manage to avoid the regions that are already fused with high-confidence proposal features to be covered again by holding an accumulated confidence map $\mathbf{C}$, which is initialized with all-zero value and updated along with recurrently computed attention maps.
Formally, the pseudo-code is provided in Algorithm~\ref{alg:ffb}.

\begin{algorithm}
    \SetKwInOut{Input}{Input}\SetKwInOut{Output}{Output}
    \Input{Backbone feature $\mathbf{F}_b$, screentone feature of multiple proposals $\{\hat{\mathbf{F}}_s^l\}_{l=0}^L$, accumulated confidence map $\mathbf{C}$, spatial attention block $\Phi$, residual block $R$}
    \Output{Fused feature $\mathbf{F}$}
    \BlankLine
    $\mathbf{F} \leftarrow \mathbf{F}_b$\;
    $\mathbf{C}^0 \leftarrow \textbf{0}$\;
    \For{$l\leftarrow 1$ \KwTo $L$}
    {
        $\mathbf{M}^l \leftarrow \Phi(\hat{\mathbf{F}}_s^{l-1}, \mathbf{F})$\;
        $\mathbf{M}_{norm}^l \leftarrow \mathbf{M}^l*(\textbf{1}-\mathbf{C}^{l-1})$\;
        $\mathbf{F} \leftarrow \mathbf{F}*(\textbf{1}-\mathbf{M}_{norm}^l)+\hat{\mathbf{F}}_s^l*\mathbf{M}_{norm}^l$\;
        $\mathbf{C}^l \leftarrow \mathbf{C}^{l-1}+\mathbf{M}^l*(\textbf{1}-\mathbf{C}^{l-1})$\;
        $\mathbf{F} \leftarrow R(\mathbf{F})$\;
    }
  \caption{Recurrent Proposal Selection Module (RPSM)}
  \label{alg:ffb}
\end{algorithm}


\subsection{Loss function}
\label{sec:loss}
We formulate four loss terms to train our network, namely translation-invariant screentone loss $\mathcal{L}_{\rm sis}$, ScreenVAE map loss $\mathcal{L}_{\rm scr}$, attention loss $\mathcal{L}_{\rm atn}$, and adversarial loss $\mathcal{L}_{\rm adv}$. 

\textbf{Translation-invariant screentone loss.}
As demonstrated in Fig.~\ref{fig:misalign}, screentones are translation-insensitive to Human Vision System (HVS), which indicates multiple acceptable candidates exist for given ground truth. To allow such ambiguity in similarity measurement, we propose the translation-invariant screentone loss $\mathcal{L}_{\rm sis}$. Specifically, for a region $p$ of the generated manga image $\tilde{\img}$, we first search an optimal offset $\delta \in \mathbb{R}^2$ for the provided ground-truth screentone template $\tilde{\img}_{t(p)}^s$, which is of the same size with $\tilde{\img}$ but only filled with the ground-truth screentone type $t(p)$, so that the pixel-wise difference of the screentones is minimized in this region. Apparently, the measurement for each region is performed separately because their optimal offset is different.
Formally, the translation-invariant screentone loss is then defined as:
\begin{equation}
    \mathcal{L}_{\rm sis} = \sum_p \min_{\delta \in w}\|\mathbf{M}_p\odot(\tilde{\img}-{\rm Shift}(\tilde{\img}_{t(p)}^s,\delta))\|_2,
\end{equation}
where $w$ is a window of $11\times 11$ and ${\rm Shift}(\cdot,\delta)$ denotes to shift an image with offset $\delta$. 
Meanwhile, as there are some screentones with width frequency, a small window may not cover all solutions while a larger window may make the model hard to converge. 
Thus, we adopt a multiscale scheme to progressively find the optimal offset. We introduce a half-size resampled version of the generated output and ground truth image, and then obtain the offset $\delta^{\prime}$ for the resampled version. The offset $\delta^{\prime}$ can be used to refine the ground truth image ${\tilde{\img}_{t(p)}^s}{}^{\prime}={\rm Shift}(\tilde{\img}_{t(p)}^s, \delta^{\prime})$, which are further used to calculate translation-invariant screentone loss.  
In particular, we use a 1-by-1 window size for the original scale to preserve the screentone appearance.

\textbf{ScreenVAE map loss.}
Empirically, under the translation-invariant screentone loss only, the model may generate screentones with visually similar but spatially incoherent patterns. Such pattern inconsistency significantly hurts the visual quality. So, the ScreenVAE map loss $\mathcal{L}_{\rm scr}$ is employed to further ensure the generated manga image is filled with the same screentone types as the ground truth. 
It is formulated as the difference between the screenVAE map of generated manga and the ground-truth ScreenVAE map $\hat{\screen}$.
\begin{equation}
    \mathcal{L}_{\rm scr} = \|{\rm SVAE}(\tilde{\img})-\hat{\screen}\|_2,
\end{equation}
where ${\rm SVAE}$ denotes the pretrained ScreenVAE model~\cite{xie2020manga} that computes the ScreenVAE map from the input manga image.

\textbf{Attention loss.} 
The attention loss $\mathcal{L}_{\rm atn}$ is developed to guide the Recurrent Proposal Selection Module (RPSM) to be more decisive.
When applying the spatial attention $\{\mathbf{M}^l\}_{l=1}^L$ to proposals, each pixel value of the fused features is represented as a weighted sum of pixels in the proposals (as presented in Algorithm~\ref{alg:ffb}). Considering the extracted feature proposals $\{\hat{F}_s^l\}_{l=0}^L$ may have inconsistent screentone-translation bias, for each region, we encourage the model to choose only one of the proposals through the attention loss:
\begin{equation}
    \mathcal{L}_{\rm atn} = \sum_l \| |\mathbf{M}^l-0.5|-0.5\|
\end{equation}
where $|\cdot|$ denotes the operation to take the element-wise absolute value. 
In particular, for synthetic data, we can directly construct the ground-truth attention mask $\hat{\mathbf{M}}^l$, and reformulate the attention loss as:
\begin{equation}
    \mathcal{L}_{\rm atn} = \sum_l \|\mathbf{M}^l-\hat{\mathbf{M}}^l\|_2
\end{equation}
To be specific, $\hat{\mathbf{M}}^l$ is the overlapped regions of the resampled label map and the anchor-based sampled label map. 

\begin{figure}
    \centering
    \subfigure[Input]{\includegraphics[width=.19\linewidth]{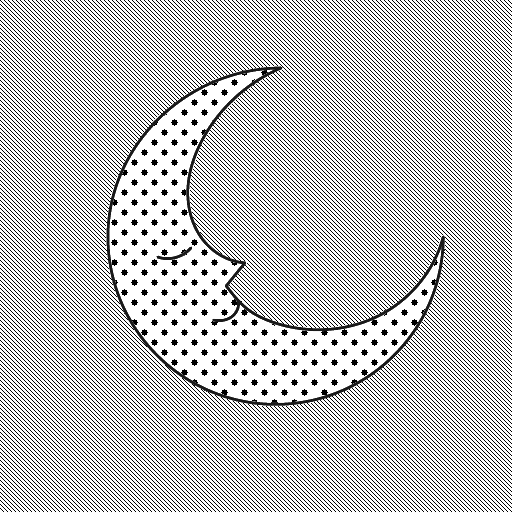}}
    \subfigure[$\mathbf{M}^1$]{\includegraphics[width=.19\linewidth]{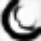}}
    \subfigure[$\mathbf{M}^2$]{\includegraphics[width=.19\linewidth]{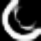}}
    \subfigure[$\mathbf{M}^4$]{\includegraphics[width=.19\linewidth]{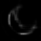}}
    \subfigure[$\mathbf{M}^8$]{\includegraphics[width=.19\linewidth]{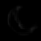}}
    \caption{Our method can generate an attention map to use appropriate proposals. (b)-(d) is the extracted attention map for each proposal under 65\% resolution.}
    \label{fig:attention}
\end{figure}


\textbf{Adversarial loss.}
The adversarial loss encourages the generated manga to follow the appearance distribution as the real-world manga, e.g., clear and binary-valued appearance, conditioned on the structural line and ScreenVAE map. We adopt Conditional GAN \cite{mirza2014conditional} to impose this constraint, with a discriminator $D_{\rm mg}$ with 5 strided downscaling blocks employed. The adversarial loss is formulated as: 
\begin{equation}
    \mathcal{L}_{\rm adv} = \sum \{\log (1-D_{\rm mg}(\tilde{\img}, \tilde{\struct}, \tilde{\screen})) + \log D_{\rm mg}(\img, \struct, \screen)\}
\end{equation}


The overall loss function for our network is defined as the weighted sum of the above terms:
\begin{equation}
    \mathcal{L} = \beta_{\rm sis}\mathcal{L}_{\rm sis}  + \beta_{\rm scr}\mathcal{L}_{\rm scr} + \beta_{\rm atn}\mathcal{L}_{\rm atn} + \beta_{\rm adv}\mathcal{L}_{\rm adv},
    \label{equ:loss}
\end{equation}
where $\beta_{\rm sis}$, $\beta_{\rm scr}$, $\beta_{\rm atn}$ and $\beta_{\rm adv}$ are the weight coefficients for different loss terms. We empirically set $\beta_{\rm sis}=10$, $\beta_{\rm scr}=100$, $\beta_{\rm atn}=5$, and $\beta_{\rm adv}=1$ in our experiments. 


%% file: results.tex
\section{Results and Discussion}
\subsection{Implementation Details}
\label{subsec:implementation}
\paragraph{Data Preparation}
We use two types of data to train our model, including synthetic manga data and real manga data. 
We manually collected 100 line drawings and 125 types of screentones. To synthesize a manga image, screentones are randomly picked and get laid onto each closed region by following the method~\cite{li2017deep}. We synthesized a total number of 6000 manga images and their corresponding per-pixel labels (i.e. screentone type). The labels are used to calculate the translation-invariant screentone loss $\mathcal{L}_{\rm sis}$ and $\mathcal{L}_{\rm atn}$. 
For the real-world data, we manually collected 20,000 screened manga of resolution 2,048$\times$ 1,536 to train our model. Note that there are no annotations for real manga data. The translation-invariant screentone loss $\mathcal{L}_{\rm sis}$ is not calculated for these data. 
All training images are cropped to $512 \times 512$ resolution during training.

\paragraph{Training Scheme}
We implemented our model in the PyTorch framework \cite{paszke2017automatic}  and trained with the loss function defined in Eq.~\ref{equ:loss}. 
The hyper-parameters in the loss function are set as following: $\beta_{\rm sis}=10$, $\beta_{\rm scr}=100$,  $\beta_{\rm atn}=5$, and $\beta_{\rm adv}=1$. 
The network weights are randomly initialized using the method of \cite{he2015delving}. We used $512 \times 512$ images for the training and a scaling factor range from 0.5 to 1.25. The Adam solver \cite{kingma2014adam} is applied to our model with a batch size of 1 and an initial learning rate of 0.0001. 
The synthetic data was first used during the training to use the appearance information of multiple proposals adaptively. Then, the whole framework was trained with both synthetic and real data to improve generalization.
We empirically found that this strategy helps to converge and improves the overall performance. 

\subsection{Comparison}
\label{subsec:comparison}
There is no existing method developed for generating retargeted manga with different resolutions. To evaluate the performance of our method, we adopt three related works for comparison: (i)\cite{tsubota2019synthesis}, first classifies the screentone types and then applied the screentone back to the line drawings; (ii)\cite{xie2020manga}, which proposes Screentone Variational AutoEncoder(ScreenVAE) to encode manga into an interpolative representation that can be used to reconstruct the input manga; and (iii) \cite{xie2021seamless}, which reuses the screentone in known area to fill up the content-missing regions by exploiting semantic correlation.
We directly use the publicly released code of \cite{tsubota2019synthesis} and \cite{xie2020manga} to generate retargeted manga. As~\cite{xie2021seamless} it is not tailored for inpainting the whole image, we make certain adaptation that enables the applicability to our manga retargeting problem. To be specific, we remove the semantic inpainting network and directly feed the resampled ScreenVAE map and structure lines, together with the original manga image, into the appearance synthesis network, which will generate a manga image of the retargeted resolution. The adapted model is retrained on our dataset, under the same training scheme as ours.

\begin{figure*}
    \centering
    \begin{minipage}[b]{\linewidth}
        \settoheight{\tempdim}{\includegraphics[width=0.19\textwidth]{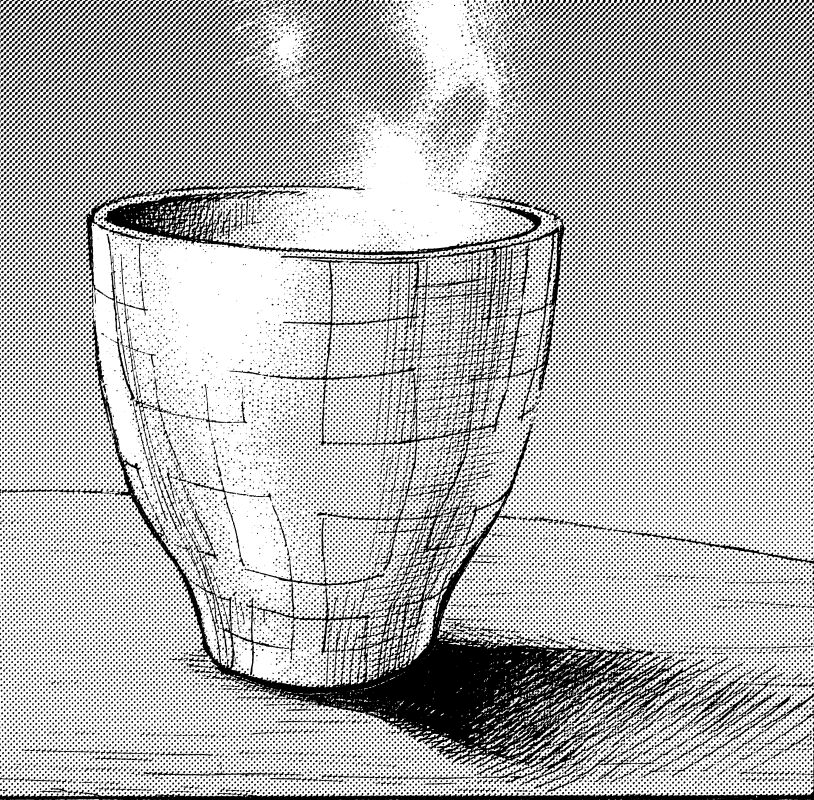}}%
        \rotatebox{90}{\makebox[\tempdim]{s=50\%}}\hfil
        \includegraphics[width=0.19\textwidth]{imgs/compare/real/bj13_009_009_orig.png}\hfil
        \includegraphics[width=0.19\textwidth]{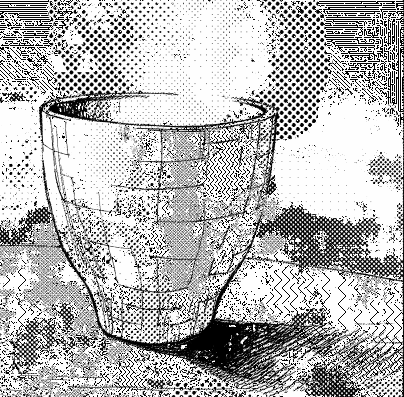}\hfil
        \includegraphics[width=0.19\textwidth]{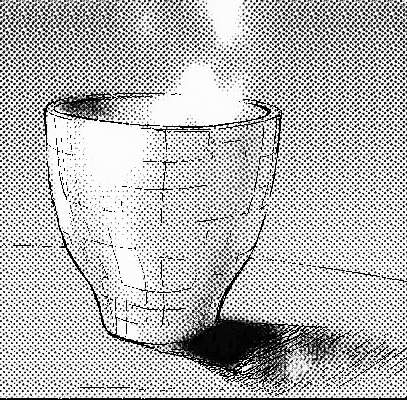}\hfil
        \includegraphics[width=0.19\textwidth]{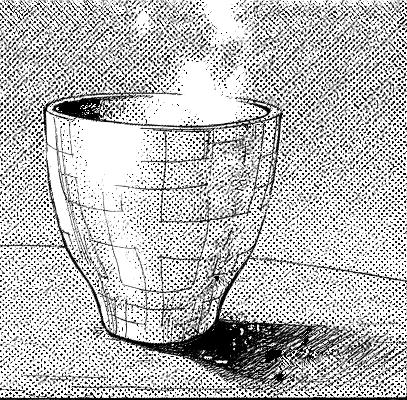}\hfil
        \includegraphics[width=0.19\textwidth]{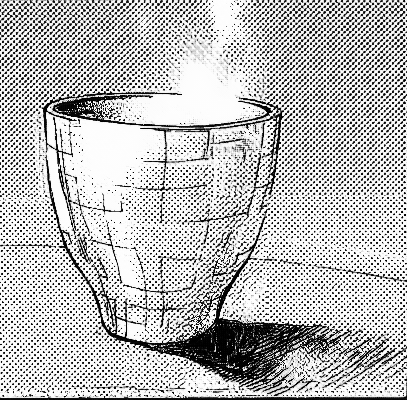}
    \end{minipage}
    \begin{minipage}[b]{\linewidth}
        \settoheight{\tempdim}{\includegraphics[width=0.19\textwidth]{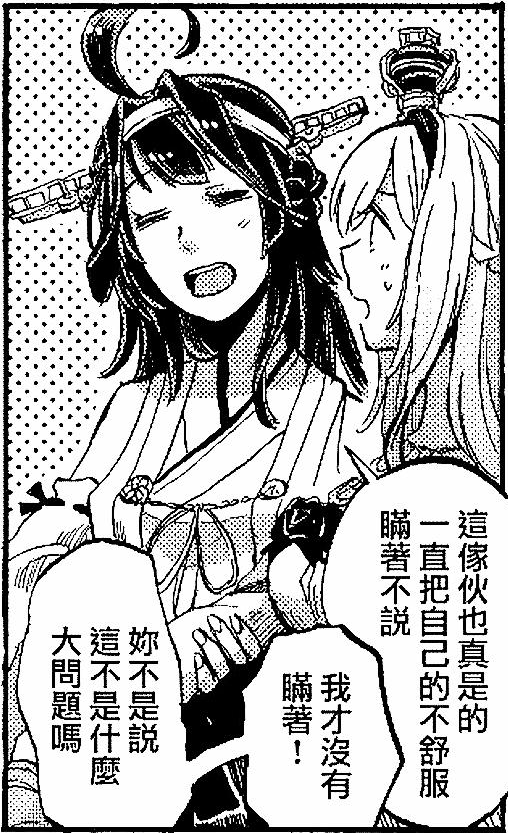}}%
        \rotatebox{90}{\makebox[\tempdim]{s=70\%}}\hfil
        \includegraphics[width=0.19\textwidth]{imgs/compare/results/16/006.png}\hfil
        \includegraphics[width=0.19\textwidth]{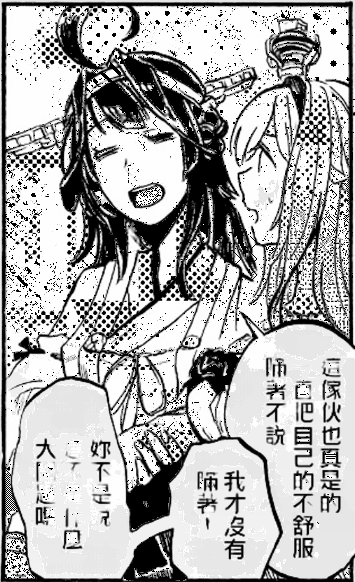}\hfil
        \includegraphics[width=0.19\textwidth]{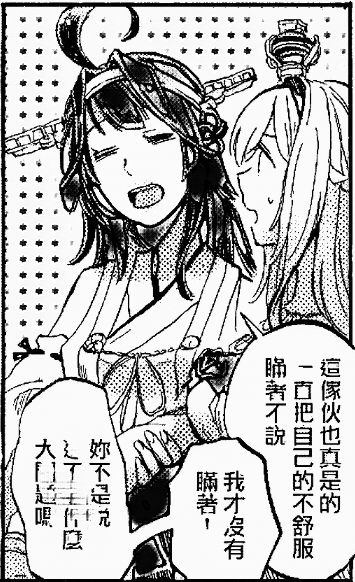}\hfil
        \includegraphics[width=0.19\textwidth]{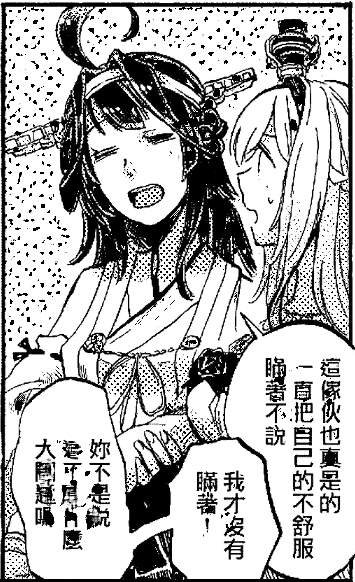}\hfil
        \includegraphics[width=0.19\textwidth]{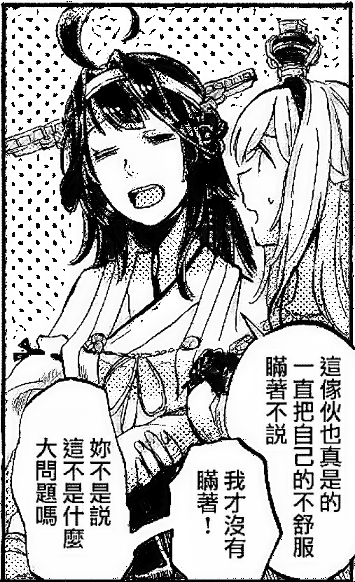}
    \end{minipage}
    \begin{minipage}[b]{\linewidth}
        \settoheight{\tempdim}{\includegraphics[width=0.19\textwidth]{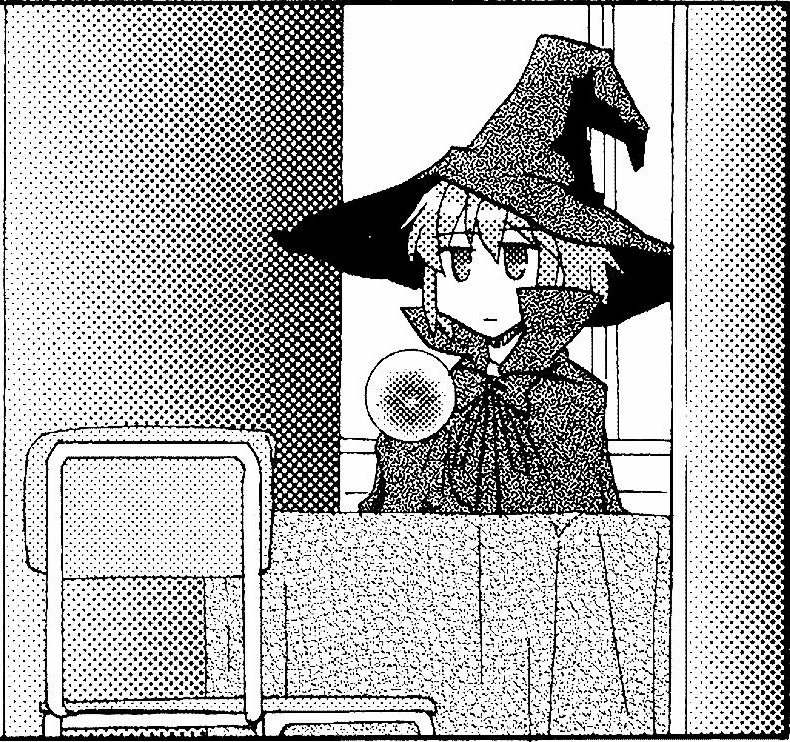}}%
        \rotatebox{90}{\makebox[\tempdim]{s=75\%}}\hfil
        \includegraphics[width=0.19\textwidth]{imgs/compare/results/3/007_orig.png}\hfil
        \includegraphics[width=0.19\textwidth]{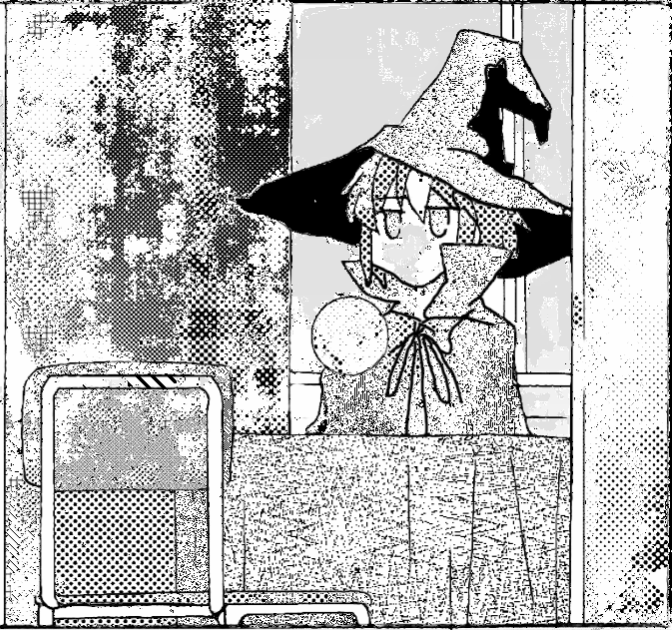}\hfil
        \includegraphics[width=0.19\textwidth]{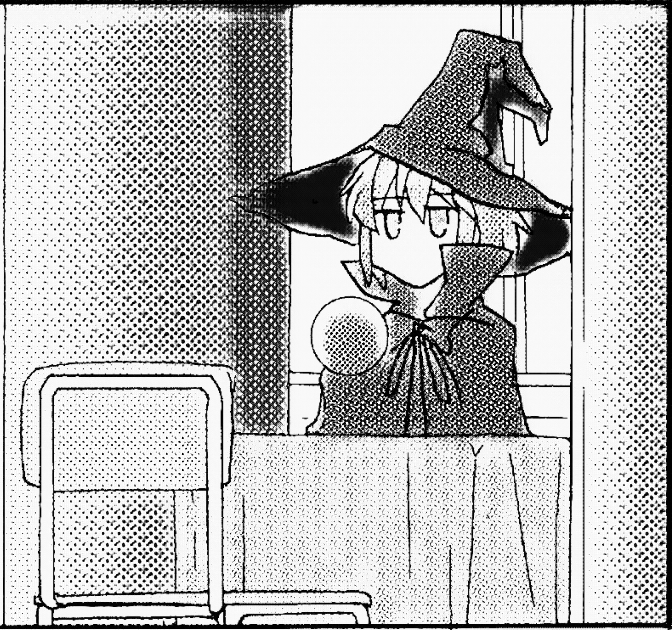}\hfil
        \includegraphics[width=0.19\textwidth]{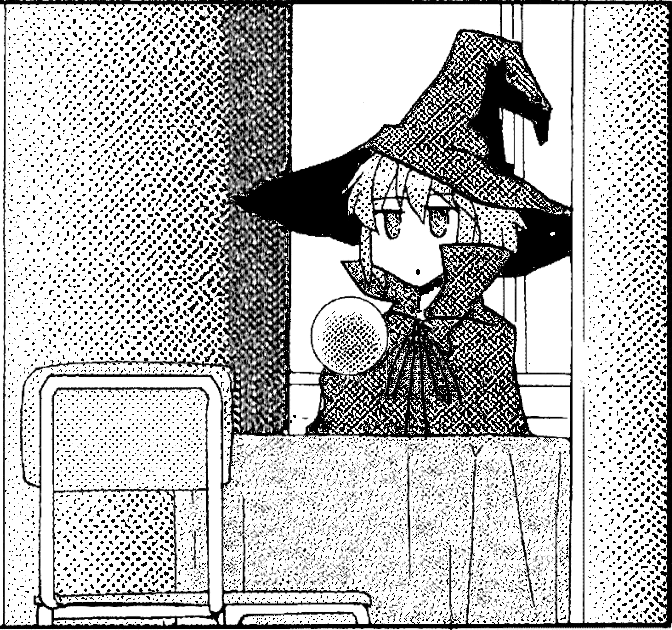}\hfil
        \includegraphics[width=0.19\textwidth]{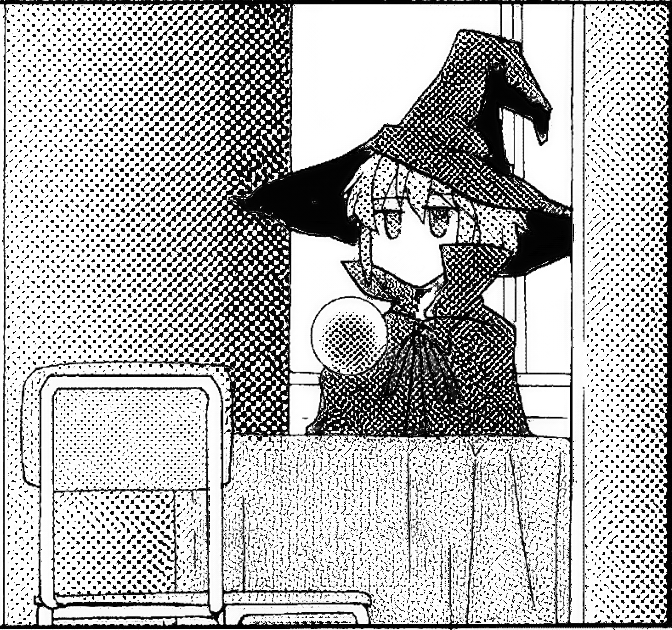}
    \end{minipage}
    \begin{minipage}[b]{\linewidth}
        \settoheight{\tempdim}{\includegraphics[width=0.19\textwidth]{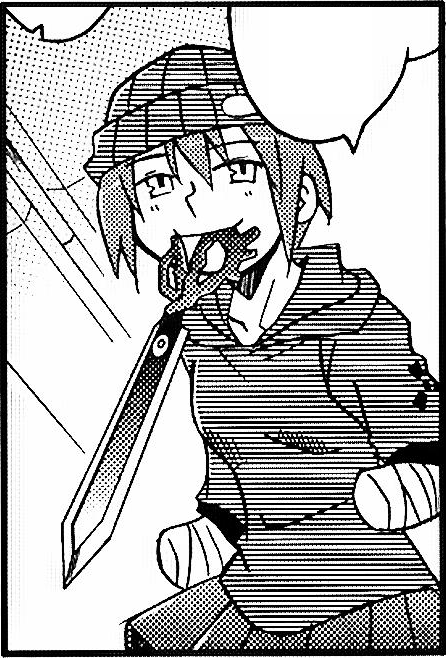}}%
        \rotatebox{90}{\makebox[\tempdim]{s=75\%}}\hfil
        \includegraphics[width=0.19\textwidth]{imgs/compare/results/6/048_orig.png}\hfil
        \includegraphics[width=0.19\textwidth]{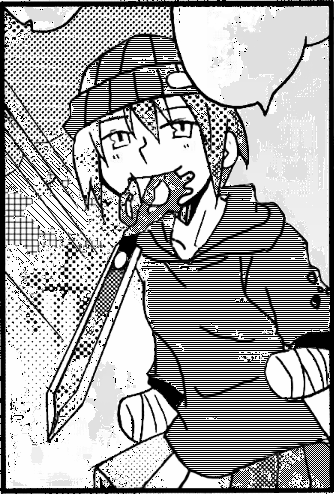}\hfil
        \includegraphics[width=0.19\textwidth]{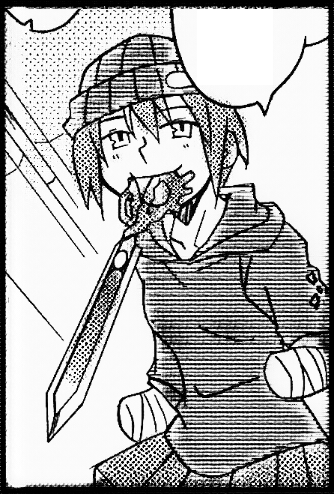}\hfil
        \includegraphics[width=0.19\textwidth]{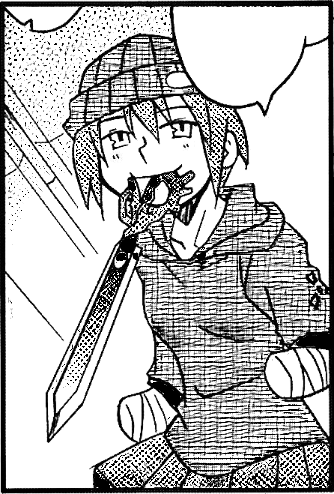}\hfil
        \includegraphics[width=0.19\textwidth]{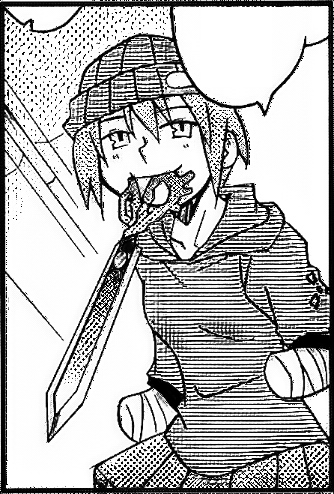}
    \end{minipage}
    \begin{minipage}[b]{\linewidth}
        \settoheight{\tempdim}{\includegraphics[width=0.19\textwidth]{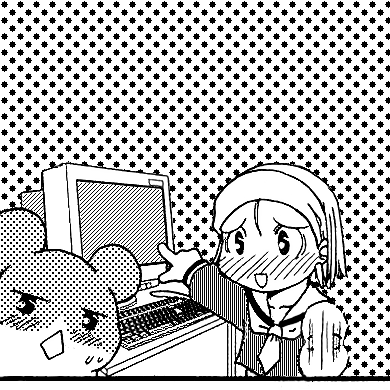}}%
        \rotatebox{90}{\makebox[\tempdim]{s=85\%}}\hfil
        \includegraphics[width=0.19\textwidth]{imgs/compare/results/31/Akuhamu_tmp0_orig.png}\hfil
        \includegraphics[width=0.19\textwidth]{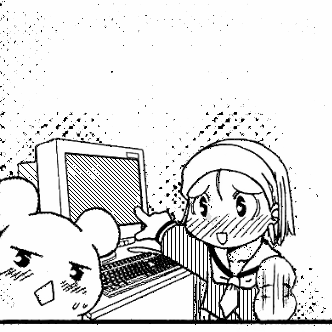}\hfil
        \includegraphics[width=0.19\textwidth]{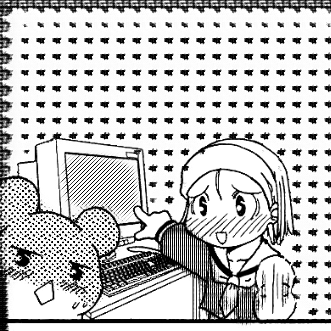}\hfil
        \includegraphics[width=0.19\textwidth]{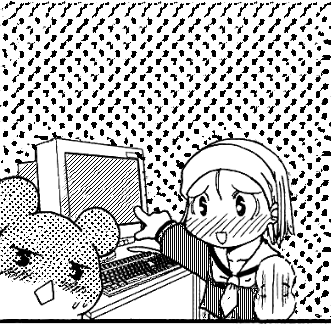}\hfil
        \includegraphics[width=0.19\textwidth]{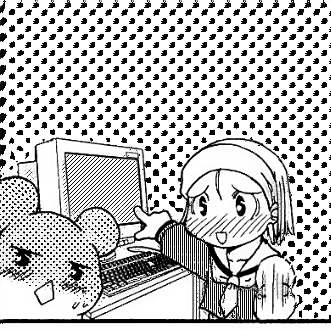}
    \end{minipage}
    \medskip
    \hspace{0.2\baselineskip}\hfil
    \makebox[0.19\textwidth]{(a) Original manga}\hfil
    \makebox[0.19\textwidth]{(b) \cite{tsubota2019synthesis}}\hfil
    \makebox[0.19\textwidth]{(c) \cite{xie2020manga}}\hfil
    \makebox[0.19\textwidth]{(d) \cite{xie2021seamless}}\hfil
    \makebox[0.19\textwidth]{(e) Ours}
    \caption{Comparison with existing methods on real-world cases.}
    \label{fig:comparison}
\end{figure*}

Fig.~\ref{fig:comparison} shows the visual results produced by the competitors mentioned above and our methods. When retargeting the manga to other resolutions, we need to guarantee consistent screentones on corresponding regions. We can see that, in general, our method shows plausible results with homogenous screentones over regions to show a great visual impression. 
The results generated by \cite{tsubota2019synthesis} may generate inconsistent screentones over the same screentone region if the screentones cannot match well with the screentones in predefined sets. Meanwhile, their method often cannot generate the same labels for the same screentone regions. 
We also compared the manga filling style translation method proposed by Xie et al.~\shortcite{xie2020manga}. However, their method also may not generate screentones in the original manga, as shown in Fig.~\ref{fig:comparison}. Besides, blurred screentones and boundary artifacts may also be generated. 
As for \cite{xie2021seamless}, the target region is filled with pixel-wisely sampled screentone features, i.e., from surrounding/reference screentone features, which can not well guarantee the screentone fidelity, especially the spatial coherence of screentone patterns. This phenomenon gets even severer when the image resolution is changed. Instead, our method just utilizes the encoded feature map (from resampled semantic representation) as a query to retrieve patch-wise proposals that retain the intactness of the input screentone features. This is the key to that our method reconstructs high-fidelity screentones whatever the retargeted resolution changes.

Besides the qualitative comparison, we quantitatively evaluate the visual quality of the results produced by various methods. We first generate 500 synthetic manga images with screentone labels. These images are further retargeted to fit resolution range from 50\% and 125\% by these methods. We adopt 3 metrics to evaluate the quality of these retargeted images, including Peak Signal-to-Noise Ratio (PSNR), Structural Similarity Index Measure (SSIM)~\cite{wang2004image}, and Learned Perceptual Image Patch Similarity (LPIPS)~\cite{zhang2018unreasonable}. We evaluate the LPIPS metric based on the VGG16 ~\cite{simonyan2014very} model. Specifically, these metrics all have limitations in evaluating the retargeted images with multiple acceptable results. These metrics may over-penalize a slight shift of screentone as they are alignment-sensitive. Thus, we measure each result with the best-matched ground truth by aligning each screentones separately. The quantitative evaluation is listed in Table~\ref{tab:comparison}. We can see that our method quantitatively outperforms other methods. 

\begin{table}
    \caption{Comparison with existing methods on synthetic data.}
    \label{tab:comparison}
    \centering
    \begin{tabular}{l||c|c|c}\hline
        Methods & PSNR & SSIM & LPIPS \\  \hline
        \cite{tsubota2019synthesis} & 6.5611 & 0.3191 & 0.4036 \\  \hline
        \cite{xie2020manga}  & 7.7150 & 0.3120 & 0.3946 \\ \hline
        \cite{xie2021seamless}  & 7.7817 & 0.3364 & 0.3589 \\ \hline
        Ours  & \textbf{9.7034} & \textbf{0.5545} & \textbf{0.2582}\\ \hline
    \end{tabular}
\end{table}


\begin{figure*}[!t]
    \centering
    \subfigure[Original image]{\includegraphics[width=.2\linewidth]{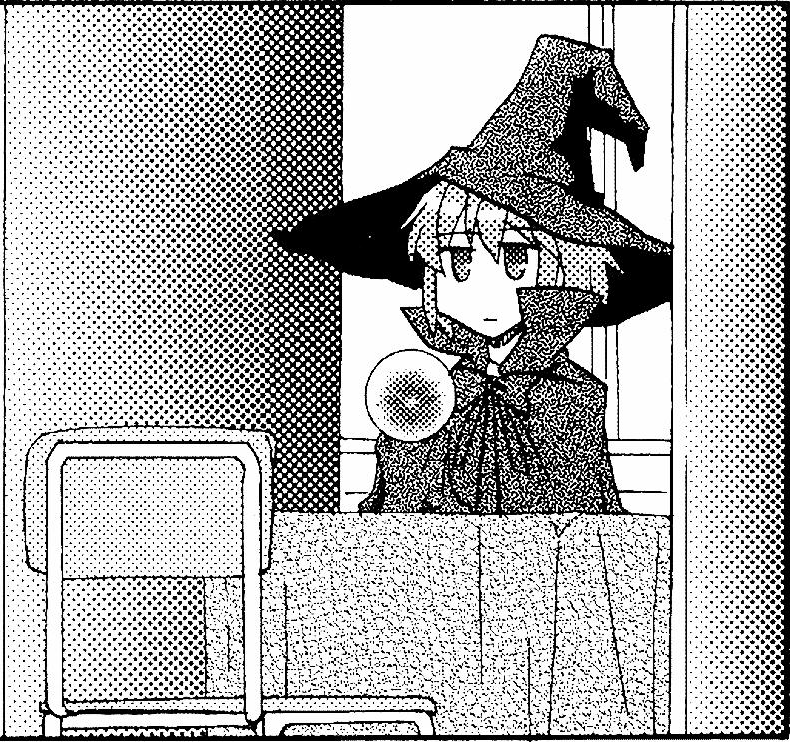}}
    \subfigure[w/o $\mathcal{L}_{\rm sis}$]{
    \includegraphics[width=.15\linewidth]{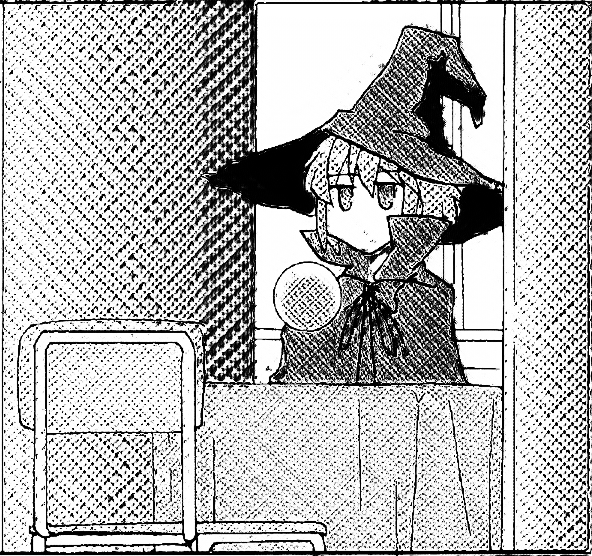}}
    \subfigure[w/o $\mathcal{L}_{\rm scr}$]{
    \includegraphics[width=.15\linewidth]{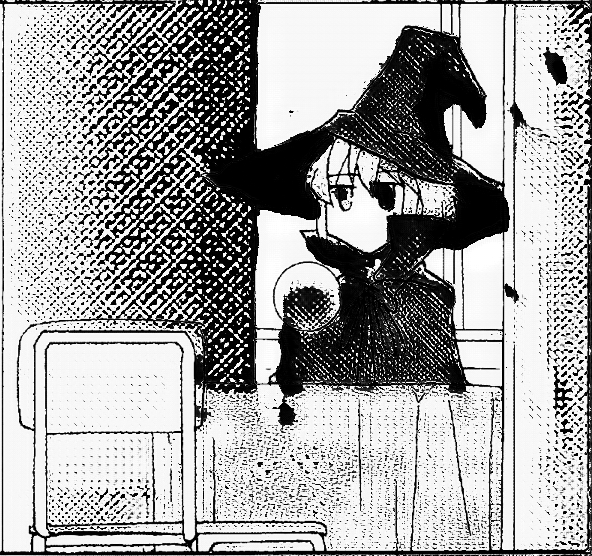}}
    \subfigure[w/o $\mathcal{L}_{\rm atn}$]{
    \includegraphics[width=.15\linewidth]{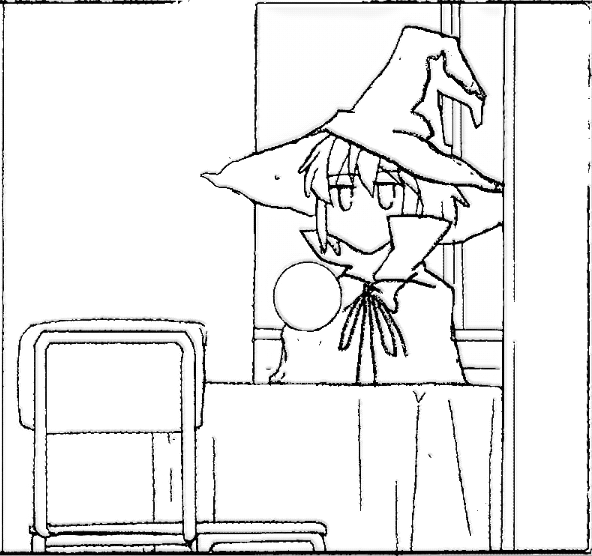}}
    \subfigure[w/o $\mathcal{L}_{\rm adv}$]{
    \includegraphics[width=.15\linewidth]{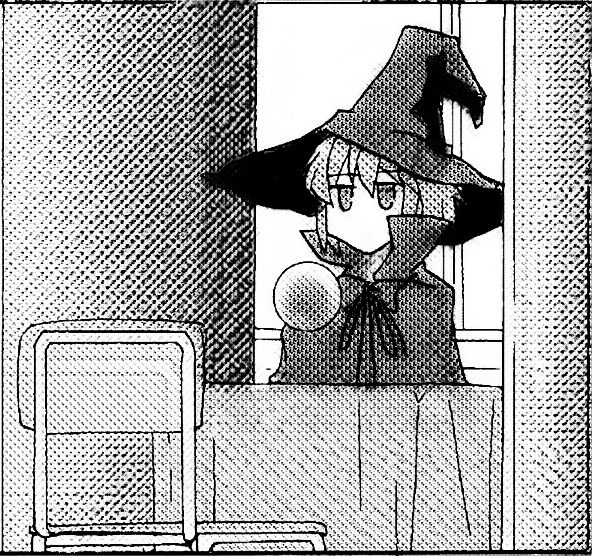}}
    \subfigure[Ours]{
    \includegraphics[width=.15\linewidth]{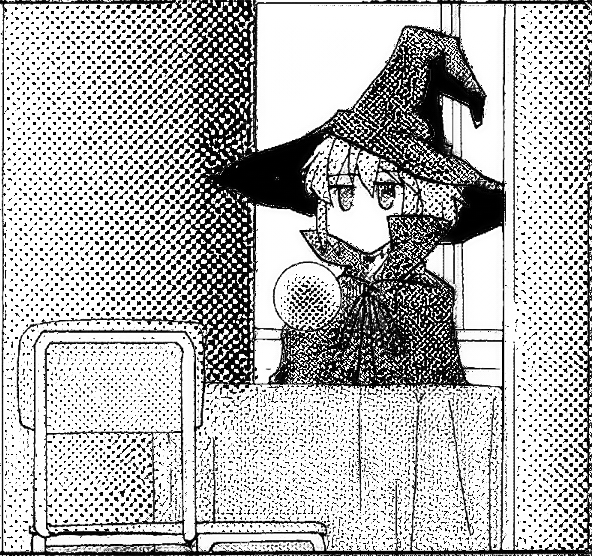}}
    \caption{Comparison with individual loss term under 75\% resolution.}
    \label{fig:ablation}
\end{figure*}

\subsection{Ablation Study}
\label{subsec:ablation}
To verify the effectiveness of each individual loss term, we conduct ablation studies for each module by visually and quantitatively comparing their generated outputs. 
Fig.~\ref{fig:ablation} shows the generated manga image of the trained model without individual loss terms. 
Without the translation-invariant screentone loss $\mathcal{L}_{\rm sis}$, the original periodic information of screentones may not be restored (Fig.~\ref{fig:ablation}(b)). The screentone types may not be reproduced without the ScreenVAE map loss $\mathcal{L}_{\rm scr}$ (Fig.~\ref{fig:ablation}(c)). The attention loss $\mathcal{L}_{\rm atn}$ is essential for model training and it helps the retargeted features to be replaced with the appropriate proposals (Fig.~\ref{fig:ablation}(d)). Without the adversarial loss $\mathcal{L}_{\rm adv}$, the network may fail to generate tidy and binarized results (Fig.~\ref{fig:ablation}(e)). 
In comparison, the combined loss can help the network generate clear and aligned screentones (Fig.~\ref{fig:ablation}(f)). The quantitative evaluation in Table~\ref{tab:ablation} also shows that the combined loss quantitatively outperforms the others variants of our method. 

\begin{table}
    \caption{Comparison without individual loss terms on synthetic data.}
    \label{tab:ablation}
    \centering
    \begin{tabular}{l||c|c|c}\hline
        Model variant & PSNR & SSIM & LPIPS \\\hline
        w/o $\mathcal{L}_{\rm sis}$  & 8.2430 & 0.3551 & 0.3950 \\\hline
        w/o $\mathcal{L}_{\rm scr}$  & 8.4949 & 0.4631 & 0.3455 \\\hline
        w/o $\mathcal{L}_{\rm atn}$  & 5.8920 & 0.1861 & 0.6134 \\\hline
        w/o $\mathcal{L}_{\rm adv}$  & \textbf{9.7680} & 0.4990 & 0.3707 \\\hline
        Ours  & 9.7034 & \textbf{0.5545} & \textbf{0.2582}\\ \hline
    \end{tabular}
\end{table}

\subsection{Limitations}
\label{subsec:limitation}
\begin{figure}
    \centering
    \begin{minipage}[b]{\linewidth}
        \settoheight{\tempdim}{\includegraphics[width=0.48\linewidth]{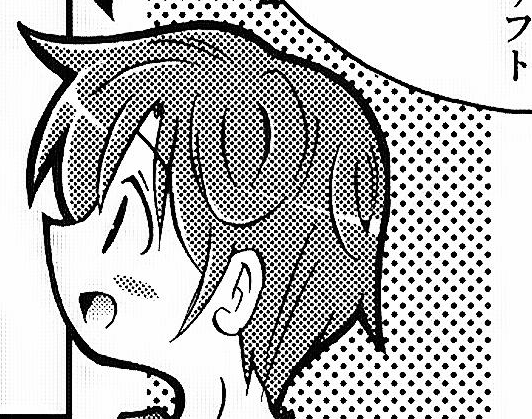}}%
        \rotatebox{90}{\makebox[\tempdim]{Original manga}}\hfil
        \includegraphics[width=0.45\linewidth]{imgs/limitation/039_orig.png}\hfil
        \includegraphics[width=0.48\linewidth]{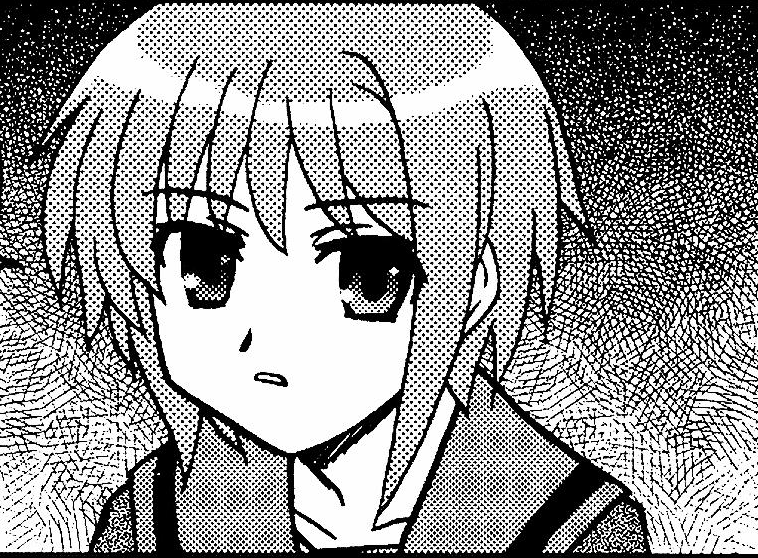}
    \end{minipage}
    \begin{minipage}[b]{\linewidth}
        \settoheight{\tempdim}{\includegraphics[width=0.48\linewidth]{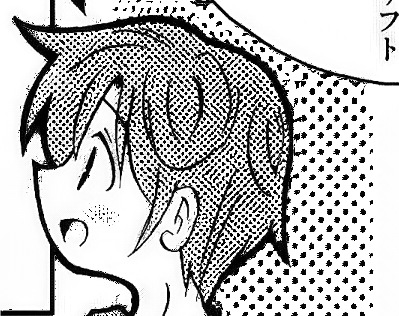}}%
        \rotatebox{90}{\makebox[\tempdim]{Retarted manga}}\hfil
        \includegraphics[width=0.45\linewidth]{imgs/limitation/039_ref.png}\hfil
        \includegraphics[width=0.48\linewidth]{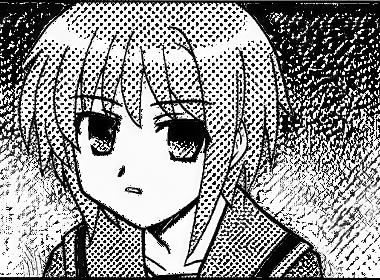}
    \end{minipage}
    \medskip
    \hspace{0.2\baselineskip}\hfil
    \makebox[0.45\linewidth]{(a) Rescaling ratio: 75\%}\hfil
    \makebox[0.45\linewidth]{(b) Rescaling ratio: 50\%}\hfil
    \caption{Limitation illustration. (a) Large-scale patterns along narrow structures. (b) Irregular patterns in the background.}
    \label{fig:limitation}
\end{figure}

Our framework still has some limitations. 
Firstly, our model may fail to generate screentones for some tiny or narrow regions, which are hard to extract periodic information. For example, in Fig.~\ref{fig:limitation}(a), the screentones between the hair and the balloon failed to generate. 
In addition, our proposed method might generate distorted patterns for irregular screentones. An example can be found in Fig.~\ref{fig:limitation}(b), and the irregular patterns in the background cannot be appropriately retargeted. This is because the candidates for irregular patterns cannot be aligned with translation. 

%% file: conclusion.tex
\section{conclusion}
In this paper, we made the first attempt to tackle the screentone-preserved manga retargeting task and obtain plausible results. We simplify the screentone synthesis problem as an anchor-based proposals selection and rearrangement problem. Hierarchical anchors-based proposal sampling is proposed to generate aliasing-free screentone proposals which are then adaptively fused to generate retargeted images through a Recurrent Proposal Selection Module. Besides, as there can be multiple solutions for manga retargeting, we propose a translation-invariant screentone loss to tolerant the misalignment of multiple possible solutions. Both the visual comparison and the quantitative experiments show the superiority of our proposed method.